\expandafter\def\csname ver@fixltx2e.sty\endcsname{}
\documentclass{elsarticle}

\usepackage{url}
\usepackage{microtype}

\usepackage{dblfloatfix}
\usepackage[utf8]{inputenc}
\usepackage{natbib}
\usepackage{amsmath}
\usepackage{amsthm}
\usepackage{amssymb}
\usepackage{amsfonts}
\usepackage{bm}
\usepackage{siunitx}
\usepackage{epstopdf}          
\usepackage{flushend}
\usepackage{parskip}
\usepackage[hyperfootnotes=true,colorlinks=true,linkcolor=blue, bookmarks=true,citecolor={red},urlcolor=cyan]{hyperref}
\usepackage[pagewise]{lineno}
\usepackage{framed} 
\usepackage{multicol} 
\usepackage{tabularx} 
\usepackage{longtable} 
\usepackage[ruled,vlined]{algorithm2e}

\journal{}
\usepackage{amssymb}

\usepackage[flushright]{rotating}
\usepackage{multirow} 

\usepackage{subcaption} 
\usepackage[printonlyused]{acronym} 
\usepackage{mathtools} 
\usepackage{comment}

\usepackage[dvipsnames]{xcolor}
\usepackage{tikz}
\usetikzlibrary{fit}
\usetikzlibrary{positioning}
\usetikzlibrary{shapes,arrows}

\definecolor{bluePolimi}{RGB}{22, 44, 80}
\definecolor{lightBluePolimi}{RGB}{91, 122, 172}
\definecolor{greenPolimi}{RGB}{0, 110, 0}
\definecolor{redPolimi}{RGB}{180, 0, 0}

\newcolumntype{M}[1]{>{\centering\arraybackslash}m{#1}}
\usepackage{booktabs} 
\usepackage{comment}
\usepackage{hyperref}

\begin{document}

\newtheorem{theo}{Theorem}
\theoremstyle{definition}
\newtheorem{obs}{Remark}
\newtheorem{Def}{Definition} 
\begin{frontmatter}

\title{Application of parametric Shallow Recurrent Decoder Network to magnetohydrodynamic flows in liquid metal blankets of fusion reactors}

\author[First]{Matteo Lo Verso}
\author[First]{Carolina Introini}
\author[First]{Eric Cervi}
\author[Second]{Laura Savoldi}
\author[Third]{J. Nathan Kutz}
\author[Fourth,First,cor1] {Antonio Cammi}

\cortext[cor1]{Corresponding author. Email address: antonio.cammi@polimi.it}

\address[First]{Department of Energy, Politecnico di Milano, Milano, 20133, Italy}
\address[Second]{MATHEP Group, Dept. of Energy "Galileo Ferraris", Politecnico di Torino, Torino, Italy}
\address[Third]{Autodesk Research, 6 Agar Street, London UK}
\address[Fourth] {Department of Mechanical and Nuclear Engineering \& Emirates Nuclear Technology Center, Khalifa University, Abu Dhabi, 127788, United Arab Emirates}

\begin{abstract}
Magnetohydrodynamic (MHD) phenomena play a pivotal role in the design and operation of nuclear fusion systems, where electrically conducting fluids (such as liquid metals or molten salts employed in reactor blankets) interact with magnetic fields of varying intensity and orientation, thereby influencing the resulting flow dynamics. The numerical solution of MHD models entails the resolution of highly nonlinear, multiphysics systems of equations, which can become computationally demanding, particularly in multi-query, parametric, or real-time contexts.
This study investigates a fully data-driven framework for MHD state reconstruction that integrates dimensionality reduction through Singular Value Decomposition (SVD) with the SHallow REcurrent Decoder (SHRED), a neural network architecture designed to reconstruct the full spatio-temporal state from sparse time-series measurements of selected observables, including previously unseen parametric configurations. 
The SHRED methodology is applied to a three-dimensional geometry representative of a portion of a WCLL blanket cell, in which lead–lithium flows around a water-cooled tube. Multiple magnetic field configurations are examined, including constant toroidal fields, combined toroidal–poloidal fields, and time-dependent magnetic fields. Across all considered scenarios, SHRED achieves high reconstruction accuracy, robustness, and strong generalization to magnetic field intensities, orientations, and temporal evolutions not encountered during training. Notably, in the presence of time-varying magnetic fields, the framework accurately infers the temporal evolution of the magnetic field itself using temperature measurements alone. Overall, the findings identify SHRED as a computationally efficient, data-driven, and flexible approach for MHD state reconstruction, with significant potential for real-time monitoring, diagnostics and control in fusion reactor blanket systems.

\end{abstract}


\end{frontmatter}

\section{Introduction}\label{sec: intro}

Magnetohydrodynamics (MHD) studies the dynamics of electrically conducting fluids under the influence of magnetic fields \cite{freidberg2014ideal}. This theory is particularly exploited in nuclear fusion research, particularly in magnetic confinement fusion (MCF). Notably, thermonuclear plasmas can be modelled as conducting fluids confined by strong magnetic fields. Moreover MHD theory also can be exploited for describing the electrically conducting fluids foreseen in the blankets of many tokamaks, such as molten salts \cite{ferrero2023impact} or liquid metals \cite{molokov2007liquid}. Indeed in tokamak configurations, residual magnetic field lines from the plasma chamber can penetrate the blanket, interacting with the conducting fluids and strongly affecting their flow regime. Consequently, understanding MHD effects in blankets is essential for reactor design, not only under nominal operating conditions but also during transients such as plasma disruptions. Given the current state of MCF development, numerical investigations of these phenomena are essential.

MHD models consist of strongly coupled, nonlinear partial differential equations that integrate fluid dynamics and magnetic fields within a multiphysics framework \cite{biskamp1997nonlinear}. Their numerical solution is computationally demanding, and the flow responses are highly sensitive to the specific profile of the magnetic field, in terms of intensity and orientation \cite{buhler2007liquid}. Simulating all possible configurations is therefore prohibitive, particularly for real-time control applications, where full-order models (FOMs) must predict unforeseen conditions but often require excessive computational time. This challenge is common in multiphysics scenarios governed by nonlinear,  multiphysics equations. In this context, Reduced Order Modelling (ROM) \cite{lassila2014model, rozza_model_2020} has emerged as a promising approach to mitigate computational costs in parametric studies. ROMs consists in surrogate models constructed from high-fidelity datasets, projecting the system dynamics onto a low-dimensional manifold spanned by dominant spatial modes via techniques such as Singular Value Decomposition (SVD). These surrogate models retain the essential physics while drastically reducing computational effort, enabling very fast predictions even for previously unseen parameter combinations. Moreover, ROM techniques can facilitate rapid parametric exploration and uncertainty quantification analyses, making them particularly attractive for control-oriented and design applications in fusion devices. While data-driven ROM approaches are well established in computational physics domains, including nuclear fission \cite{riva2025data, riva2025real, riva2024impact, riva2024multi, cammi2024data}, their application to MHD (and fusion more broadly) has only recently begun to emerge \cite{RoyKutz2018_Plasma, kaptanoglu2020characterizing, kaptanoglu2021physics}, remaining especially limited for scenarios involving electrically conducting liquid metals \cite{loverso2024application, LOVERSO2025115080, loverso_NURETH}.

In parallel, the fusion community has increasingly adopted Machine Learning (ML) and Artificial Intelligence (AI) techniques for real-time control, monitoring, and digital twin applications \cite{battye2025digital}. Deep learning architectures have been successfully applied to plasma control \cite{degrave2022magnetic, wang2025learning}, instability mitigation \cite{seo2024avoiding}, and profile regulation \cite{jalalvand2021real} in tokamaks, demonstrating impressive capabilities in learning complex nonlinear dynamics and enabling rapid predictions. However, purely data-driven approaches typically require large training datasets and extensive training times, which can be prohibitive when high-fidelity simulations are costly or experimental data are scarce, noisy, or difficult to acquire. These limitations are particularly relevant in MHD scenarios involving liquid metal flows in fusion blankets, where generating extensive datasets under realistic operating conditions is the principal challenge.

A promising alternative is to integrate ML within a reduced, physically informed framework. By compressing the system’s dimensionality using ROM techniques, learning can be performed in a low-dimensional latent space, significantly reducing the dimensionality of training data and computational effort while preserving essential physical characteristics of the MHD dynamics. This approach bridges physics-based modelling and data-driven methods and is particularly suited for fusion MHD applications. In addition, it facilitates the integration of experimental measurements with prior model knowledge, offering advantages over conventional data assimilation methods, which are often limited by high computational costs.

Within this context, the present study explores a combination of SVD and ML for accurate and reliable MHD state reconstruction in parametric scenarios. The chosen architecture is the SHallow REcurrent Decoder (SHRED) \cite{williams2024sensing, faraji2025shallow}, which maps sparse trajectories of measured observables to the full high-dimensional state, indirectly estimating unmeasured quantities. Through a recurrent unit followed by a shallow decoder, SHRED efficiently captures the spatio-temporal dynamics of the system, even when trained with a limited number of sensors. Importantly, it generalizes across different parameter values, making it suitable for reconstructing flows under a range of magnetic field profiles. \newline
To date, SHRED has been successfully applied to a diverse set of physical systems \cite{williams2024sensing, faraji2025shallow, tomasetto2025reduced, moen2025mapping}, consistently exhibiting strong accuracy and robust generalization. Within nuclear engineering, this strategy has been previously employed for state estimation in fission reactor scenarios \cite{riva2025constrained, riva2025robust, riva2025towards, introini2025models}. At the present day, applications of SHRED to MHD physics of electrically conducting liquids involved in fusion reactors remain very limited in the literature. In a previous work \cite{loverso2026SHRED}, the authors applied SHRED to a lead-lithium MHD flow in a two-dimensional channel with steps and thermal gradients, training the network over a broad range of magnetic field intensities. The results showed that, starting from temperature measurements at only three randomly placed sensors, SHRED accurately reconstructed the full multiphysics state (temperature, velocity, and pressure) across the entire domain, successfully generalizing to magnetic field intensities not seen during the training phase. Furthermore, the model demonstrated remarkable robustness to sensor placement, exhibiting negligible variability in reconstruction accuracy across multiple randomized sensor configurations.  While the test case in the previous paper featured a relatively complex physics, it relied on a simplified geometry that is not directly representative of any realistic fusion blanket configuration. In the present work, the objective is to further assess the capabilities of SHRED in a scenario characterized by an equally complex physical behaviour, but within a geometry that is more representative and directly relatable to blanket-relevant configurations. The selected test case consists in a three-dimensional portion of a blanket elementary cell, involving lead–lithium with realistic thermophysical properties and subjected to different magnetic field configurations with intensities, orientations and temporal profiles relevant to fusion reactor conditions. 

The structure of the present paper is organized as follows. Section \ref{sec: SHRED} introduces the SHRED architecture. Section \ref{sec: Numerical results} presents the MHD model and discusses the key numerical results. Finally, Section \ref{sec: Conclusions} summarizes the main conclusions of the present study, along with some future perspectives.

\section{SHallow REcurrent Decoder}\label{sec: SHRED}
\begin{figure}[htbp]
    \centering
    \includegraphics[width=1.0\linewidth]{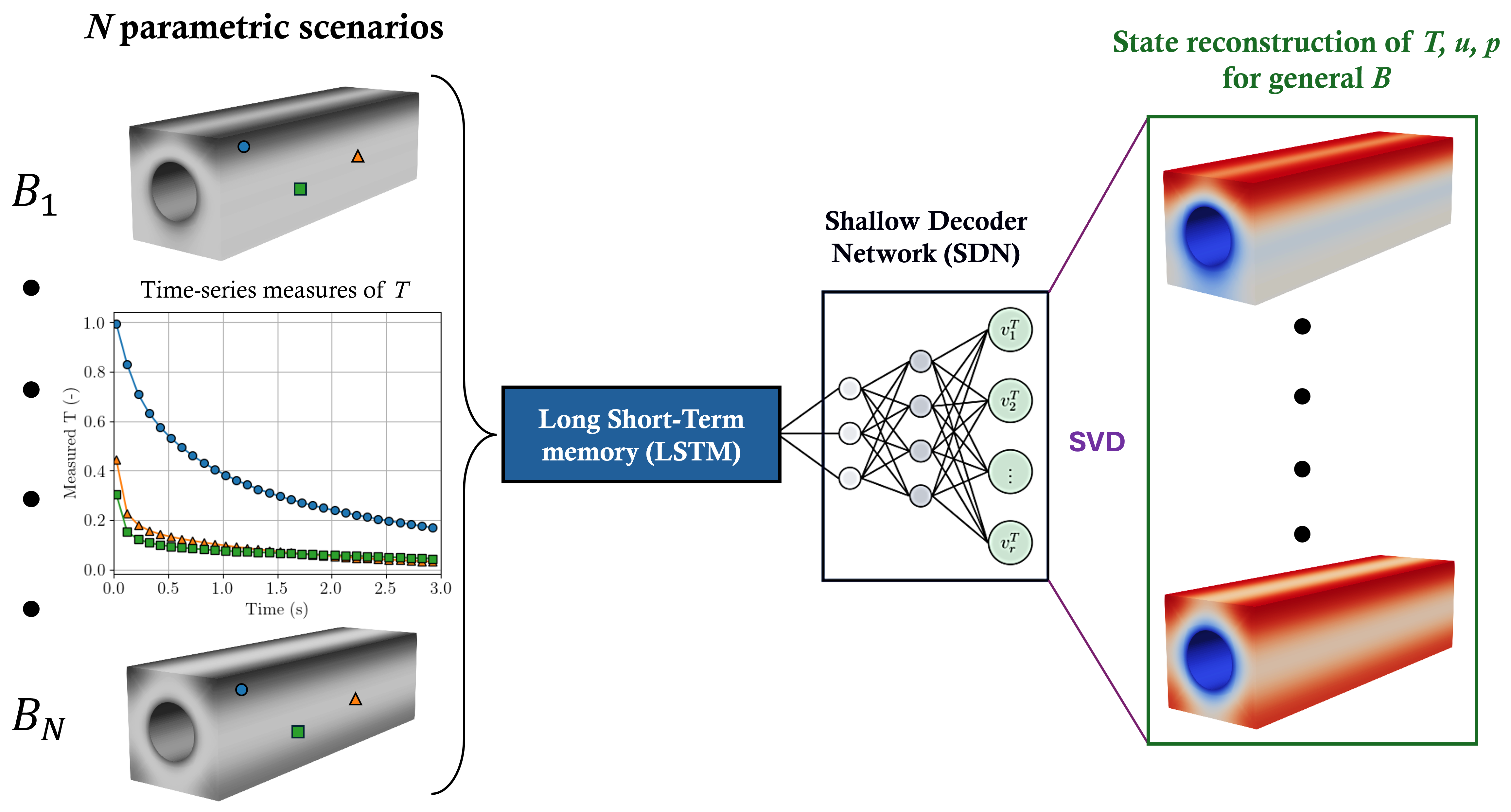}
    \caption{Schematic representation of the SHRED architecture applied to the considered three-dimensional test case. After compressing the original dataset via SVD, three sensors record the local temporal evolution of the temperature field. The resulting time-series signals are encoded into a latent space through a Long Short-Term Memory (LSTM) network. Subsequently, a Shallow Decoder Network (SDN) maps the latent representation onto a compressed approximation of the full set of spatio-temporal field variables. Finally, the SVD basis is used to reconstruct the corresponding full-order state back-projcting its compressed representation.}
    \label{fig: SHRED3D_architecture}
\end{figure}

The SHallow REcurrent Decoder (SHRED) is a recently proposed data-driven machine learning framework, first introduced in \cite{williams2024sensing}, for state estimation and forecasting of complex dynamical systems from sparse time-series observations. Its baseline architecture combines a Long Short-Term Memory (LSTM) network \cite{LSTM2016long}, which captures temporal dependencies in the latent dynamics, with a Shallow Decoder Network (SDN) \cite{SDN20shallow}, responsible for the nonlinear mapping from the latent space to the high-dimensional physical space.
In the present study, a compressed variant of SHRED is adopted. Prior to training, the dataset is reduced via Singular Value Decomposition (SVD), leading to a substantial decrease in the number of input features. The integration of SVD within the SHRED pipeline has been shown to markedly improve computational efficiency by lowering the dimensionality of the learning problem \cite{tomasetto2025reduced}, enabling training even on standard personal computer. The architecture of the SHRED network employed in this work is illustrated in Figure  \ref{fig: SHRED3D_architecture}.

First of all, the architecture learns the temporal evolution of the system trajectories in accordance with Takens’ embedding theorem \cite{takens2006detecting}, which states that the dynamics of a high-dimensional system can be reconstructed from time-delayed observations of a limited number of variables. In this framework, the LSTM captures the temporal dependencies and nonlinear correlations embedded in the sensor measurements over time. The Shallow Decoder Network (SDN) subsequently maps the learned latent trajectories onto the reconstructed space, where the SVD basis is used to decompress the latent representation and recover the full-order state of the system.

This architecture presents several advantages over conventional data-driven surrogate modelling techniques. First, SHRED has been shown to achieve accurate state reconstruction with a remarkably small number of sensors (typically three) beyond which the reconstruction error tends to plateau \cite{tomasetto2025reduced}. This feature makes it particularly suitable for scenarios where measurements are limited or sensor deployment is costly. Moreover, SHRED is largely agnostic to sensor placement, consistently delivering accurate reconstructions even with randomly distributed sensors \cite{williams2024sensing, loverso2026SHRED}. As a result, sensors can be installed in locations that are practically accessible, without the need for complex optimization procedures. This characteristic is especially beneficial in fusion environments, where geometric constraints, high temperatures, and radiation may severely restrict sensor positioning. An additional advantage lies in SHRED’s ability to infer multiphysics quantities from measurements of a single observable, exploiting learned correlations among strongly coupled variables. In tokamak systems, for instance, temperature measurements are generally more accessible than other quantities such as flow velocity or neutron flux. By leveraging correlations captured during training, SHRED enables the estimation of otherwise unmeasured variables from the most readily available signals. Compared with many alternative AI approaches based on neural networks, SHRED can be trained directly on compressed data representations, significantly reducing memory requirements and computational costs, and allowing training on standard personal computers. Furthermore, it requires minimal hyperparameter tuning, as the same architectural configuration has proven effective across diverse physical systems \cite{williams2024sensing}. \newline
A further key strength of SHRED, particularly relevant in nuclear engineering applications, is its solid mathematical grounding. The methodology builds upon Takens’ embedding theorem \cite{takens2006detecting} and can be interpreted as a data-driven extension of classical separation of variables concepts. Owing to its shallow architecture, SHRED involves a very limited number of trainable parameters ( typically fewer than $1000$) in contrast to many deep learning models that generally rely on millions of parameters. This reduced complexity enhances interpretability, promotes physical insight into the learned dynamics, and increases confidence in its deployment in safety-critical nuclear scenarios. All these features make SHRED an excellent candidate for state reconstruction in complex physics. 

In its original formulation, SHRED was developed for single-parameter scenarios \cite{williams2024sensing, faraji2025shallow, riva2025robust}, focusing on reconstructing system dynamics under fixed physical conditions. Nevertheless, the architecture can be readily extended to parametric problems, as demonstrated in \cite{tomasetto2025reduced, riva2025towards}. This adaptability arises from SHRED’s inherent design: because the LSTM operates on lagged time-series sequences, the architecture can naturally handle multiple trajectories corresponding to varying parameter values. In this parametric context, a physical parameter $\mu$ can be incorporated either as an additional input, if its value is known, or as a target output when parameter estimation is desired.

The logic of the algorithm is explained in the following. For each parameter realization $\boldsymbol{\mu}_p$, the snapshot matrix is defined as $\mathbb{X}^{\boldsymbol{\mu}_p} \in \mathbb{R}^{\mathcal{N}_h \times N_t}$, where $\mathcal{N}_h$ denotes the number of mesh cells (i.e. the number of spatial degrees of freedom) and $N_t$ the number of stored time steps. This matrix is then compressed via Singular Value Decomposition (SVD), through the reduced basis $\mathbb{U}^{\boldsymbol{\mu}_p} \in \mathbb{R}^{\mathcal{N}_h \times r}$ of rank $r$. The associated latent representation is obtained as $$
\mathbb{V}^{\boldsymbol{\mu}_p}=\left(\mathbb{U}^{\boldsymbol{\mu}_p}\right)^T \mathbb{X}^{\boldsymbol{\mu}_p} \in \mathbb{R}^{r \times N_t}$$
where $\mathbb{V}^{\boldsymbol{\mu}_p}$ contains the temporal coefficients encoding the dynamics corresponding to the parameter value $\mu_p$, and serves as training data for SHRED. In case of parametric dataset, however, it is necessary to construct a global reduced basis for spanning the entire parametric space, thereby capturing the most representative physical features across different parameter instances. To this end, the complete dataset is assembled as 
\begin{equation}    \mathbb{X}=\left[\mathbb{X}^{\boldsymbol{\mu}_1}\left|\mathbb{X}^{\boldsymbol{\mu}_2}\right| \ldots \mid \mathbb{X}^{\boldsymbol{\mu}_{N_p}}\right]
\end{equation}
where $N_p$ denotes the total number of sampled parameter values.
In the present work, the parameter of interest is represented by the magnetic field profile, which plays a key role in determining the dynamics of the lead–lithium flow within the fusion reactor blanket \cite{buhler2007liquid, muller2001magnetofluiddynamics}. Specifically, configurations characterized by markedly different magnetic fields are considered, varying not only in intensity but also in orientation and temporal evolution. Extending SHRED to such a diverse set of magnetic field profiles enables the assessment of its capability to reconstruct MHD effects across substantially different physical regimes. Demonstrating accurate performance under these heterogeneous conditions provides strong evidence of the generality and robustness of SHRED with respect to arbitrary magnetic field configurations.

The SHRED architecture has been implemented in Python using the PyTorch library, building upon the original code of \cite{williams2024sensing}. In the adopted architecture, both the LSTM and the SDN components consist of two hidden layers. The LSTM hidden layers contain $64$ neurons each and a lags of $30$ time instants, while the SDN is composed of two hidden layers with $350$ and $400$ neurons, respectively.

\section{Numerical Results} \label{sec: Numerical results}
\begin{figure}[htbp]
    \centering
    \includegraphics[width=1.0\linewidth]{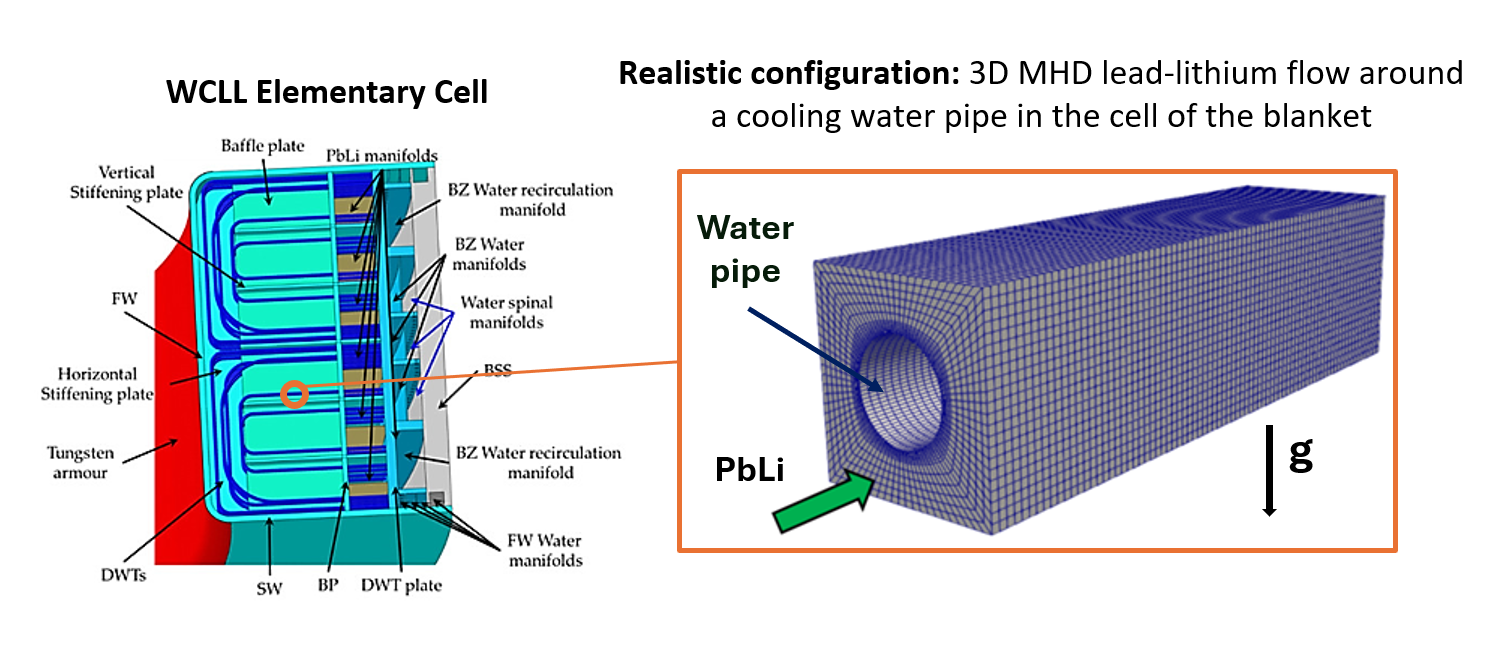}
    \caption{Visual sketch of the selected model geometry. The image of the elementary cell on the left is taken from \cite{arena2021demo}.}
    \label{fig: 3d_geometry}
\end{figure}

This work investigates the applicability of the SHRED to a configuration representative of a portion of a blanket elementary cell, involving lead–lithium with realistic thermophysical properties and subjected to magnetic fields with intensities and orientations relevant to fusion reactor conditions. 
As physical context, the Water-Cooled Lead-Lithium (WCLL) blanket design proposed for DEMO can be considered. The elementary cell of the WCLL blanket consists of a confined breeding zone module, structurally defined by horizontal and vertical stiffening plates, in which lead–lithium flows through a guided radial–poloidal–radial path and is cooled by arrays of water tubes. Since this work represents the first application of SHRED to a three-dimensional MHD scenario involving lead–lithium flows in blanket-relevant conditions, the geometrical configuration is intentionally simplified by focusing on a single cooling water tube and on the lead–lithium flow in its immediate vicinity. The computational domain consists of a three-dimensional rectangular region of length $L$ and square cross-section of side $a$, filled with flowing lead–lithium at temperature $T_0$. A cylindrical cavity of radius $r$ is embedded at the centre, representing the space occupied by the water tube. The adoption of such a simplified configuration is a common approach in the literature to understand the fundamental physical mechanisms governing lead–lithium flows in elementary blanket cells \cite{lyu20253d, koehly2019design, buhler2024liquid}. The selected model geometry is represented in Figure \ref{fig: 3d_geometry}.

The cooling pipe is modeled by imposing a temperature $T_{pipe} < T_0$  on the cylindrical surface. The external boundaries of the domain do not correspond to physical walls; instead, they represent artificial boundaries of a subdomain, allowing the fluid to cross them, as the simulated region corresponds only to a portion of the space surrounding the tube. In order to represent a fully developed flow condition, cyclic boundary conditions are applied at the inlet and outlet for all physical fields. At the initial time, the flow is considered with uniform velocity $\mathbf{u}_0$ in the axial direction. The pipe surface is considered no-slip and perfectly electrically conducting. 

The  resulting MHD model for the considered compressible, visco-resistive MHD flow is the following:
\begin{equation}
    \left\{\begin{array}{ll}
    \dfrac{\partial \rho}{\partial t} +  \nabla \cdot \left( \rho \mathbf{u}\right) = 0 & \text{in } \Omega, \, t > 0 \\[5pt]
    \dfrac{\partial (\rho \mathbf{u}) }{\partial t} + \nabla \cdot \left( \rho \mathbf{u} \otimes \mathbf{u} \right) = - \nabla p + \nabla \cdot \boldsymbol{\tau} + \rho \mathbf{g} + \left( \dfrac{1}{\mu_B} \nabla \times  \mathbf{B}\right) \times \mathbf{B} & \text{in } \Omega, \, t > 0\\[5pt]
    \dfrac{\partial (\rho c T) }{\partial t} + \nabla \cdot (\rho c T  \mathbf{u})  =   \kappa \Delta T + \dfrac{1}{\sigma \mu_B^2}|\nabla \times \boldsymbol{B}|^2 & \text{in } \Omega, \, t > 0\\[5pt]
    \dfrac{\partial{\mathbf{B}}}{\partial{t}} = \nabla \times (\mathbf{u} \times \mathbf{B}) + \dfrac{1}{ \sigma \mu_0}  \Delta \mathbf{B} & \text{in } \Omega, \, t > 0 \\[5pt]
    \nabla \cdot \mathbf{B} = 0 & \text{in } \Omega, \, t > 0\\
    \rho = \rho_0 \left(1 - \beta\left(T - T_0 \right) \right) & \text{in } \Omega, \, t > 0\\
    \boldsymbol{\tau} =\mu\left(\nabla \mathbf{u}+(\nabla \mathbf{u})^T\right)-\dfrac{2}{3} \mu(\nabla \cdot \mathbf{u} \, \mathbf{I}) & \text{in } \Omega, \, t > 0\\
    \end{array}\right.
\end{equation}
addressed with the initial conditions
\begin{equation}
    \mathbf{u} = \mathbf{u}_0 , \; T = T_0, \; \rho = \rho_0, \; p = p_0, \; \mathbf{B} = \mathbf{B}_0 \quad\text{in } \Omega, \, t = 0
\end{equation}
and boundary conditions
\begin{equation}
\label{pDMD_FOM}
    \left\{\begin{array}{ll}
    \mathbf{u}_{in} = \mathbf{u}_{out}, \; T_{in} = T_{out}, \; p_{in} = p_{out}, \; \mathbf{B}_{in} = \mathbf{B}_{out}   & \text{on } \Gamma_{inlet}-\Gamma_{outlet},   \, t > 0 \\ [5pt]
    \mathbf{u} = \mathbf{0}, \; \dfrac{\partial \mathbf{B}}{\partial \mathbf{n}} = 0, \; \dfrac{\partial p}{\partial \mathbf{n}} = 0, \; T = T_{pipe}  & \text{on } \Gamma_{pipe},  \, t > 0 \\ [10pt]
    \dfrac{\partial \mathbf{u}}{\partial \mathbf{n}} = 0, \, p = p_{ext}, \, \dfrac{\partial T} {\partial \mathbf{n}} = 0, \, \mathbf{B} = \mathbf{B}_0 & \text{on } \Gamma_{walls},  \, t > 0   \\
    \end{array}\right.
\end{equation}
where $\Omega$ is the domain and $\Gamma$ are the surfaces of the boundary. \newline
All the details on the physical and numerical parameters used are reported in Table \ref{tab: 3Dgeom_parameters}.
\begin{table}[htbp]
    \centering
    \caption{Physical and numerical parameters for the FOM.}
    \label{tab: 3Dgeom_parameters}
    \begin{tabular}{|c|c|c|c|c|c|} 
    \hline
    $\rho_0$ & $9806\ \text{kg/m}^3$ 
    & $\kappa$ & $20.93\ \text{W m}^{-1}\ \text{K}^{-1}$ 
    & $p_{\text{ext}}$ & $10^5\ \text{Pa}$ \\ \hline

    $\mu$ & $1.93\times10^{-3}\ \text{Pa}\cdot\text{s}$ 
    & $c$ & $189.5\ \text{J kg}^{-1}\text{K}^{-1}$  
    & $\mathcal{N}_h$ & $112640$ \\ \hline

    $\mu_B$ & $1.26\times10^{-6}\ \text{H/m}$ 
    & $u_0$ & $0.01\ \text{m/s}$ 
    & $L$ & $0.07\ \text{m}$ \\ \hline

    $\sigma$ & $7.82\times10^5\ \Omega^{-1}\text{m}^{-1}$ 
    & $T_0$ & $600\ \text{K}$ 
    & $a$ & $0.02\ \text{m}$ \\ \hline

    $\beta$ & $1.3\times10^{-4}\ \text{K}^{-1}$ 
    & $T_{\text{pipe}}$ & $560\ \text{K}$ 
    & $r$ & $0.005\ \text{m}$ \\ \hline
    \end{tabular}
\end{table}

\begin{figure}[htbp]
    \centering
    \includegraphics[width=1.0\linewidth]{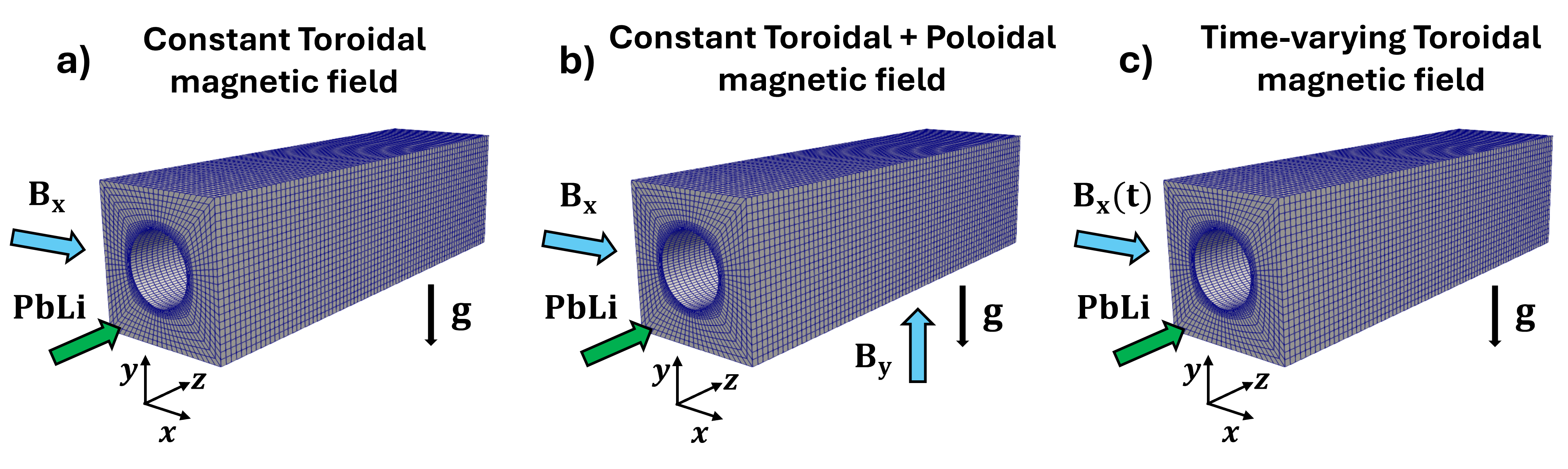}
    \caption{Overview of the selected magnetic field configurations considered in this analysis: a) constant toroidal magnetic field, b) constant toroidal toroidal + poloidal magnetic field, c) time-varying magnetic field.}
    \label{fig: B_configurations}
\end{figure}

The SHRED technique has been tested for three different scenarios involving different magnetic field configurations. The various profiles, shown in Figure \ref{fig: B_configurations}, are investigated in the following subsections. 

The snapshots needed to train SHRED were generated using the OpenFOAM MHD library \textbf{magnetoHDFoam}, developed and verified in \cite{loverso2024solver} and available on \url{https://github.com/ERMETE-Lab/MHD-magnetoHDFoam} under the MIT license. For all considered cases, simulations were run for a time of $3$ seconds, and data were stored every $0.025 \ \text{s}$, resulting in $120$ temporal instants for each case. Then data have been rescaled using the min-max formula:
\begin{equation}
    \tilde{T}= \frac{T - T_{min}}{T_{max} - T_{min}}, \quad
    \tilde{\mathbf{u}}= \frac{\mathbf{u} - \mathbf{u}_{min}}{\mathbf{u}_{max} - \mathbf{u}_{min}}, \quad\tilde{p}= \frac{p' - p'_{min}}{p'_{max} - p'_{min}}
\end{equation}
where $p' = p-\rho gh$ represents the pressure without the hydrostatic component. In the following, all variables will be considered in their normalized form, and for simplicity of notation, they will be denoted simply as $T$, $\mathbf{u}$, and $p$. \newline
For each magnetic configuration, the temperature sensors were placed at random locations within the channel, since SHRED has already demonstrated in \cite{loverso2026SHRED} that its performance is agnostic to the specific sensor positions. In all considered scenarios, the sensor positions remain fixed and are reported in Table \ref{tab: sensor_positions}.

\renewcommand{\arraystretch}{1.3}
\begin{table}[htbp]
\centering
\caption{Positions of the measurement sensors in the domain.}
\label{tab: sensor_positions}
\begin{tabular}{|c|c|c|c|}
\hline
\textbf{Sensors} & \textbf{x [m]} & \textbf{y [m]} & \textbf{z [m]} \\ \hline
S1 & 0.0070 & 0.0014 & 0.0617 \\ \hline
S2 &  0.0051 & -0.0018 & 0.0564 \\ \hline
S3 & -0.0027 & 0.0045 & 0.0066 \\
\hline
\end{tabular}
\end{table}
\renewcommand{\arraystretch}{1}

During the training stage, SHRED was provided with temperature measurements together with the compressed representations of the training and validation data, in order to learn the mapping between sparse sensor inputs and the corresponding SVD temporal coefficients. One SHRED model for each of the three magnetic configurations shown in Figure \ref{fig: B_configurations} has been trained. The training of each SHRED model required a few minutes on a standard personal computer equipped with an Intel Core i7-9800X processor. In the subsequent testing phase, SHRED receives as input only the temperature measurements from the test cases and, by exploiting the input–output relationship learned during training, reconstructs the full high-dimensional state associated with the previously unseen magnetic field intensity. The computational cost of the online reconstruction is negligible, with each reconstruction requiring less than one second. This represents a substantial computational speed-up with respect to the full-order model, whose resolution required approximately $5$ to $15$ hours (depending on the associated magnetic field profile) per simulation on a HPC.

\subsection{Constant magnetic field in toroidal direction}

\begin{figure}[htbp]
    \centering
    \includegraphics[width=1.0\linewidth]{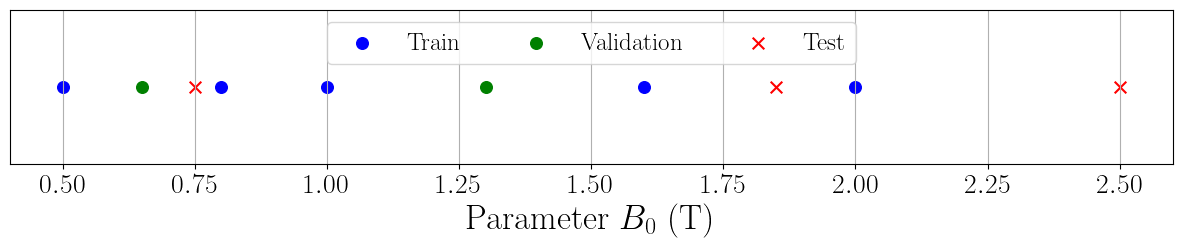}
    \caption{Subdivision of the dataset in training, validation and test snapshots.}
    \label{fig: Bx_range}
\end{figure}

First, a magnetic configuration with a toroidal magnetic field was considered, as shown in Figure \ref{fig: B_configurations}-(a). Indeed, under realistic operating conditions, the dominant component of the magnetic field in fusion reactors is the toroidal one \cite{arena2021demo, tassone2022magnetic}. The considered geometry represents only a very small domain with respect to the overall magnetic field extension. As a result, the toroidal component of the magnetic field can be approximated to act along the $x$-direction, which is transverse to the main flow direction. Accordingly, multiple snapshots were generated by considering magnetic fields with different intensities $B_x$. Figure \ref{fig: Bx_range} shows the entire generated dataset with the subdivision into training, validation, and test sets.

The training dataset spans a wide parametric range of magnetic field intensities, with $B_x \in\mathcal{B}=[0.5, 2.0]\;\text{T}$, corresponding to a window of $1.5 \;\text{T}$. Several parameter values were selected as test cases. In particular, two very different values within the training interval, namely $0.75 \;\text{T}$ and $1.85 \;\text{T}$, were chosen to assess the ability of SHRED to reconstruct flow states associated with both lower and higher magnetic field intensities and thus to retrieve a general representation, even considering different dynamics. In addition, an extrapolative test case with $2.5 \;\text{T}$, lying well outside the training range, was also considered to investigate the capability of the method to estimate the system state for parameter values outside those seen during training. The configurations investigated are subjected to magnetic fields with intensities of the same order of magnitude as those expected to be experienced by the lead–lithium within the blanket under realistic operating conditions.

\begin{figure}[htbp]
    \centering
    \includegraphics[width=1.0\linewidth]{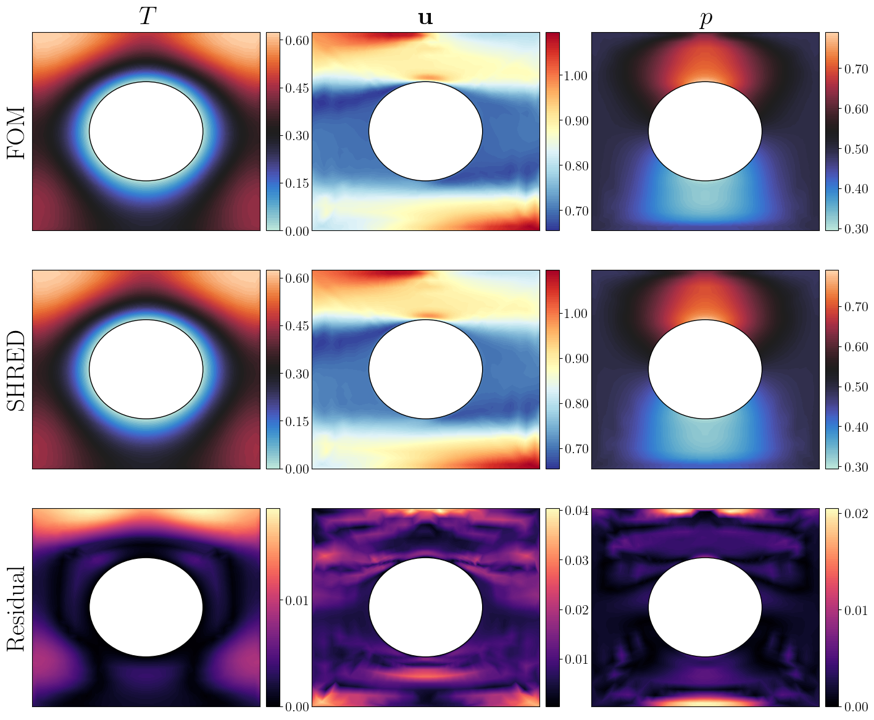}
    \caption{Results for the temperature (first column), velocity (second column) and pressure (third column) for the case with $B_x=0.75 \, T$ at time $t=2\, s$. The first row displays the reference full-order solution while the second row shows the reconstruction produced by the SHRED model. The third row reports the absolute difference between the FOM and the SHRED.}
    \label{fig: Bx075}
\end{figure}

\begin{figure}[htbp]
    \centering
    \includegraphics[width=1.0\linewidth]{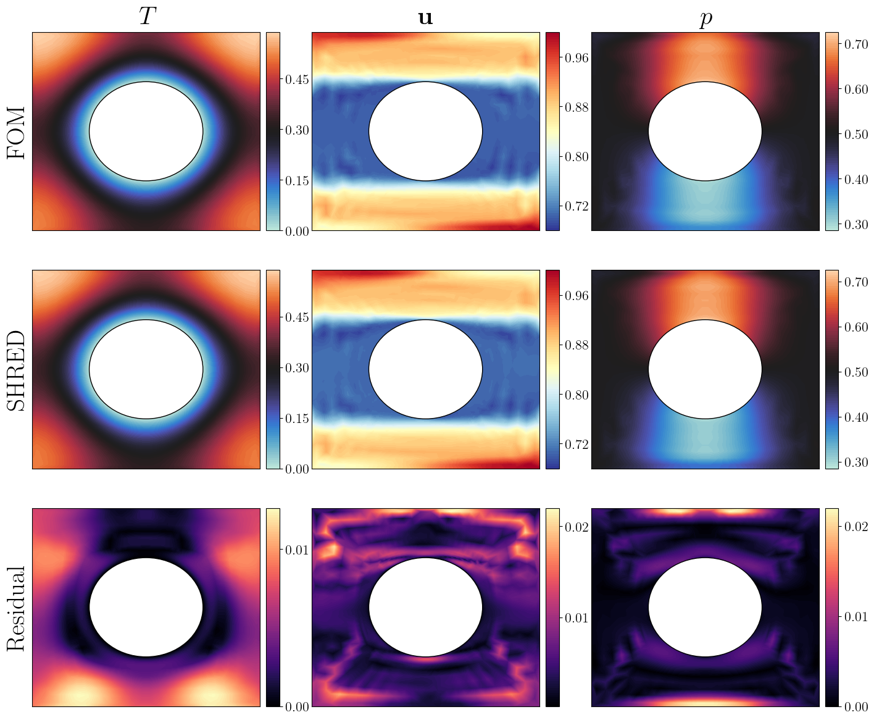}
    \caption{Results for the temperature (first column), velocity (second column) and pressure (third column) for the case with $B_x=1.85 \, T$ at time $t=2\, s$. The first row displays the reference full-order solution while the second row shows the reconstruction produced by the SHRED model. The third row reports the absolute difference between the FOM and the SHRED.}
    \label{fig: Bx185}
\end{figure}

\begin{figure}[htbp]
    \centering
    \includegraphics[width=1.0\linewidth]{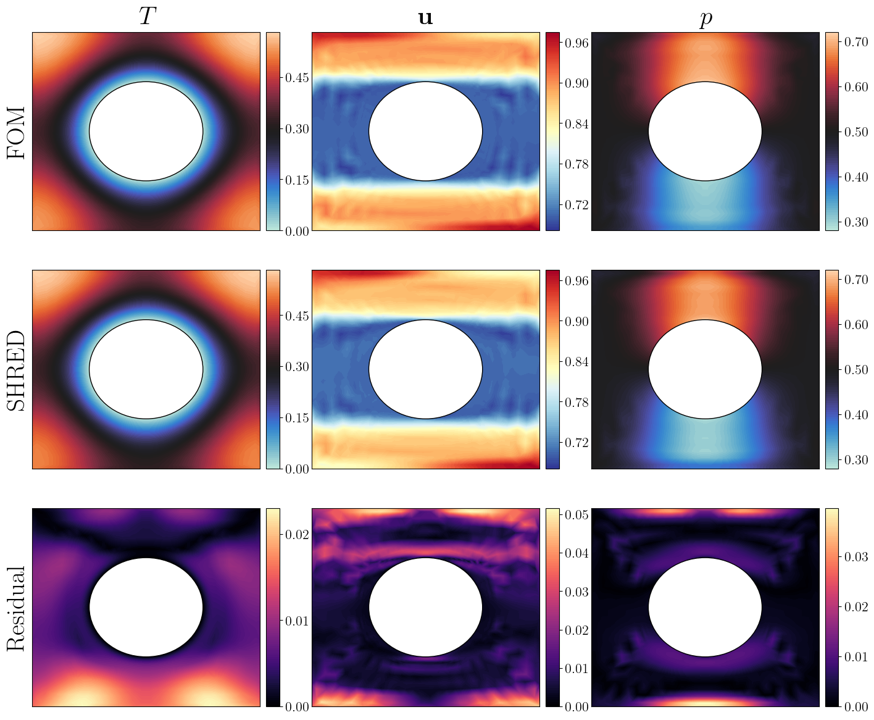}
    \caption{Results for the temperature (first column), velocity (second column) and pressure (third column) for the case with $B_x=2.5 \, T$ at time $t=2\, s$. The first row displays the reference full-order solution while the second row shows the reconstruction produced by the SHRED model. The third row reports the absolute difference between the FOM and the SHRED.}
    \label{fig: Bx250}
\end{figure}

The SHRED model was trained using the temperature measurements and the compressed representations (the first $5$ modes) of the training and validation data. In the following, the state reconstructions obtained by SHRED for the test cases are shown. Figure \ref{fig: Bx075}, \ref{fig: Bx185} and \ref{fig: Bx250} show the results obtained for the test case with magnetic field intensity $B_x = 0.75 \, \text{T}$, $B_x = 1.85 \, \text{T}$ and $B_x = 2.5 \, \text{T}$ respectively. 


The figures depict a cross-section of the square channel at mid-length, at $t=2 \ s$. The comparison includes the reference solution, corresponding to the high-fidelity model, the SHRED reconstruction, and the absolute difference between the two. For the test cases within the training parameter range, i.e. $B_x = 0.75 \, \text{T}$, $B_x = 1.85 \, \text{T}$ (Figures \ref{fig: Bx075} and \ref{fig: Bx185}), the results show that SHRED is able to reconstruct all the fields of interest with very high accuracy, relying exclusively on temperature measurements over time. Moreover, the residuals, defined as the absolute difference between the reference solution and the reconstructed one, are extremely low overall, and their maximum values reach only a few percentage points; higher residuals are confined to small, highly localized regions of the domain. It is also interesting observing the results obtained for the extrapolation test case, corresponding to the magnetic field intensity outside the training range, i.e $B_x = 2.5 \, \text{T}$ (Figure \ref{fig: Bx250}). Even in this scenario, SHRED demonstrates a remarkable capability to accurately reconstruct the flow fields. The residuals are only slightly higher than those observed for the two in-range test cases, yet they remain overall negligible and are consistently characterized by maximum values of only a few percentage points. 

The results presented so far have focused on the reconstruction of the flow fields at a fixed time instant. To further quantify the accuracy of the model throughout the full temporal horizon considered, a time-dependent quantity must be introduced. The relative $L^2$-error related to the SHRED reconstruction in the entire spatial domain over time for each field $\psi$ has been calculated as:

\begin{equation}
\label{eq: error_SHRED3D}
    \epsilon_\psi=\frac{\left\|\psi_{F O M}-\psi_{SHRED}\right\|}{\left\|\psi_{F O M}\right\|}
\end{equation}
where $\left\| \cdot \right\|$ represents the classical $L^2$-norm. 

\begin{figure}[htbp]
    \centering
    \includegraphics[width=1.0\linewidth]{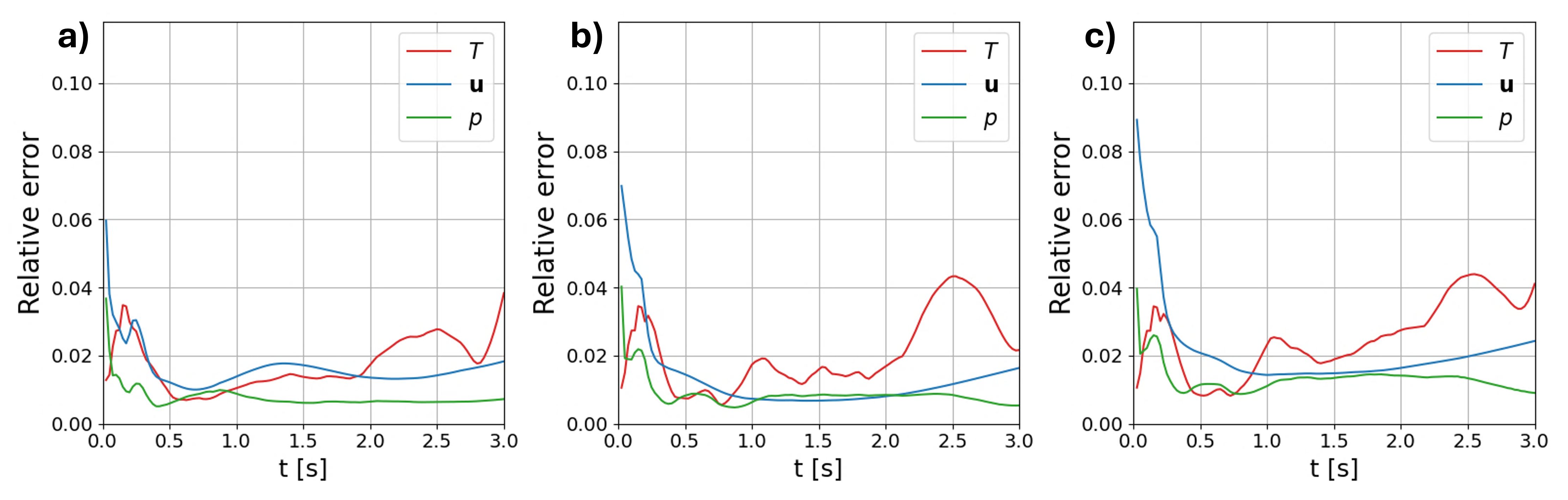}
    \caption{Temporal behavior of relative $L^2$-error of the SHRED reconstruction over time for temperature, velocity and pressure for the cases with $B_x = 0.75 \ T$ (a), $B_x = 1.85 \ T$ (b) and $B_x = 2.5 \ T$ (c).}
    \label{fig: Bxconst_error}
\end{figure}

Figure \ref{fig: Bxconst_error} shows the relative errors for the three considered test cases. In all the considered scenarios, the relative errors associated with the reconstructed physical quantities remain generally low. In particular, the velocity field exhibits higher errors at the initial time instants, although they never exceed $9\%$, and rapidly decrease as the simulation progresses, stabilizing around $2\%$ for all test cases. This behavior can be attributed to the initial transient phase of the flow evolution: the velocity field is initially uniform throughout the domain, but under the action of the magnetic field, the flow rapidly reorganizes, developing regions of higher and lower velocity. This transition phase involves fast-changing dynamics, which are intrinsically more challenging to capture accurately. Nevertheless, even during the early stages of the simulation, the velocity reconstruction error remains low and within acceptable limits. 

Concerning the temperature field, the relative error starts from very small values and exhibits a slight increase over time, while remaining consistently low and never exceeding $4\%$. This trend is related to the initial uniform temperature distribution in the domain and to the subsequent thermal evolution driven by the boundary condition imposed on the pipe wall. As the flow evolves, the cooling effect of the pipe progressively extends into the surrounding lead–lithium, leading to the growth of a colder region around the tube. Nevertheless, SHRED is able to reproduce the temperature dynamics with very good accuracy throughout the entire time window. Finally, the pressure field exhibits the lowest reconstruction error among all the considered quantities. The relative error initially reaches values of about $4\%$, but it rapidly decreases and stabilizes below $2\%$. This behavior is consistent with the physical evolution of the pressure field in the considered scenario: starting from a spatially uniform initial condition, the pressure quickly adjusts to the developing flow and subsequently maintains a fairly stable configuration over time. Generally, the reconstruction errors are consistently small for all physical fields and test cases considered, including the case corresponding to a magnetic field intensity extrapolated beyond the training parametric range. These results demonstrate that SHRED is capable of reconstructing all the tested scenarios with high accuracy, confirming its robustness and predictive capabilities in parametric MHD configurations.

\subsection{Constant magnetic field in toroidal and poloidal directions}

In realistic operating conditions, the magnetic field experienced by the liquid metal within the blanket is not purely toroidal. In addition to the dominant toroidal component, a poloidal component is also present, albeit with lower intensity. As a result, the effective magnetic field acting on the flow is more complex, combining both toroidal and poloidal contributions, with the toroidal component remaining predominant. Motivated by this physical consideration, the present analysis extends the SHRED framework to scenarios in which both magnetic field components are simultaneously taken into account. The parametric space is therefore expanded to two dimensions, with the parameters being the two magnetic field components or, equivalently, the magnitude of the resulting magnetic field and its orientation angle. 

A new set of snapshots was therefore generated by considering flow configurations subjected to a magnetic field with two components, namely a dominant toroidal component and a secondary poloidal component, as shown in Figure \ref{fig: B_configurations}-(b). Since the considered geometry represents only a very small portion with respect to the magnetic field domain, the toroidal–poloidal plane can be reasonably approximated as a Cartesian plane. Within this framework, the toroidal magnetic field component corresponds to $B_x$ while the poloidal component corresponds to $B_y$. The simulated magnetic fields were selected to reflect realistic blanket conditions, with toroidal-to-poloidal component ratios ranging approximately between $3 \div 5$. This corresponds to an orientation of the resulting magnetic field forming an angle with the toroidal plane between about $10$°$\div 20$°. The overall magnetic field intensity was varied within a realistic range, between $1  \, \text{T} \div 2 \, \text{T}$. These configurations are consistent with realistic and representative magnetic conditions experienced by liquid metal blankets in fusion reactors \cite{tassone2022magnetic}. Table~\ref{tab: BxBy_range} reports the magnetic field values considered in this study, together with the corresponding subdivision of the generated snapshots into training, validation, and test datasets. 

\renewcommand{\arraystretch}{1.2}
\begin{table}[htbp]
\centering
\caption{Magnetic field configurations and subdivision of snapshots into training, validation and test sets.}
\label{tab: BxBy_range}
\begin{tabular}{|c|c|c|c|c|}
\hline
\textbf{Snapshots} & $B_x$ & $B_y$ & $|\mathbf{B}_{tot}|$ & $\alpha$ \\
\hline
\multirow{4}{*}{Training} 
& $1.0 \, \text{T}$ & $0.2 \, \text{T}$ & $1.02 \, \text{T}$ & $11.31^\circ$ \\ \cline{2-5}
& $1.4 \, \text{T}$ & $0.35 \, \text{T}$ & $1.44 \, \text{T}$ & $14.04^\circ$ \\ \cline{2-5}
& $1.8 \, \text{T}$ & $0.55 \, \text{T}$ & $1.88 \, \text{T}$ & $16.99^\circ$ \\ \cline{2-5}
& $2.0 \, \text{T}$ & $0.7 \, \text{T}$ & $2.12 \, \text{T}$ & $19.29^\circ$ \\ \hline
Validation 
& $1.2 \, \text{T}$ & $0.3 \, \text{T}$ & $1.24 \, \text{T}$ & $14.04^\circ$ \\ \hline
Test 
& $1.6 \, \text{T}$ & $0.45 \, \text{T}$ & $1.66 \, \text{T}$ & $15.71^\circ$ \\ \hline
\end{tabular}
\end{table}

During the training phase, SHRED was trained using the temperature measurements and the compressed representations (with a rank equal to $5$) of the training and validation snapshots. The selected test case is represented by the configuration with $B_x = 1.6 \, \text{T}$ and $B_y = 0.45 \, \text{T}$, for which both parameters (namely the magnetic field intensity $|\mathbf{B}_{tot}|$ and the angle $\alpha$) take values that were not included in the training and validation sets. 

\begin{figure}[htbp]
    \centering
    \includegraphics[width=1.0\linewidth]{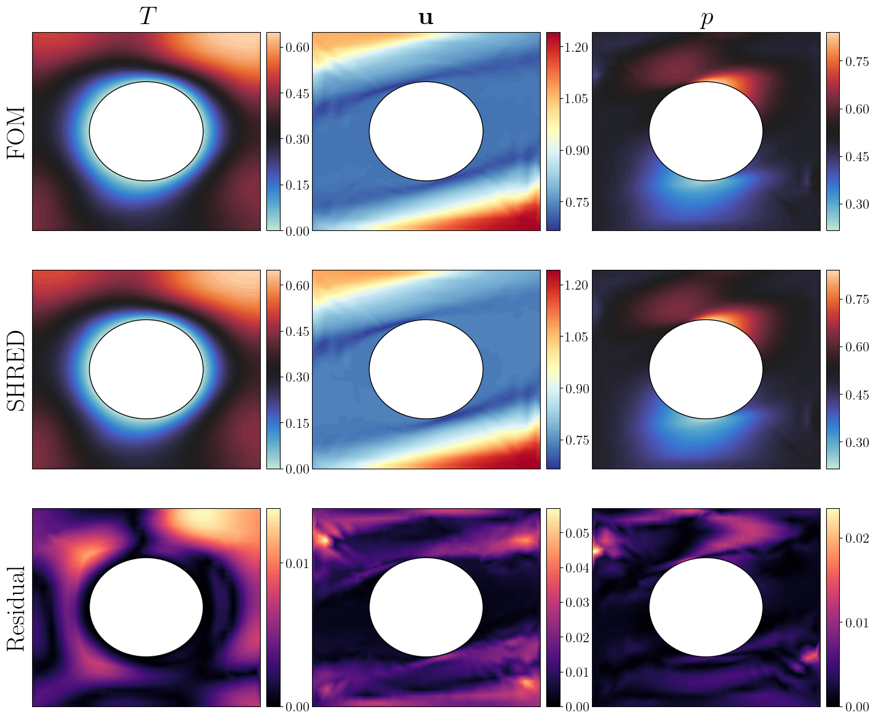}
    \caption{Results for the temperature (first column), velocity (second column) and pressure (third column) for the case with $B_x=1.60 \, \text{T}$ and $B_y=0.45 \, \text{T}$ at time $t=2\, s$. The first row displays the reference full-order solution while the second row shows the reconstruction produced by the SHRED model. The third row reports the absolute difference between the FOM and the SHRED.}
    \label{fig: Bx160By045}
\end{figure}

Figure \ref{fig: Bx160By045} shows the reconstruction obtained for the test case at the cross-section at mid-length of the geometry, at $t=2 \ s$. It can be observed that, in this configuration as well, the SHRED model is able to reconstruct the system state with a high level of accuracy. The full-order model solution and the SHRED reconstruction are extremely close for all the physical fields considered, with residuals that remain localized and reach at most a few percentage points.  It is worth emphasizing that, in this configuration, the resulting flow differs significantly from the case previously analyzed with only a constant toroidal magnetic field component. The inclusion of a poloidal component leads to an inclined flow structure, with temperature, velocity, and pressure gradients oriented along an oblique direction. This results in a more complex flow pattern compared to the case with a purely toroidal magnetic field. Nevertheless, despite the increased complexity of the dynamics, SHRED is still capable of accurately reconstructing the system evolution. 

\begin{figure}[htbp]
    \centering
    \includegraphics[width=0.7\linewidth]{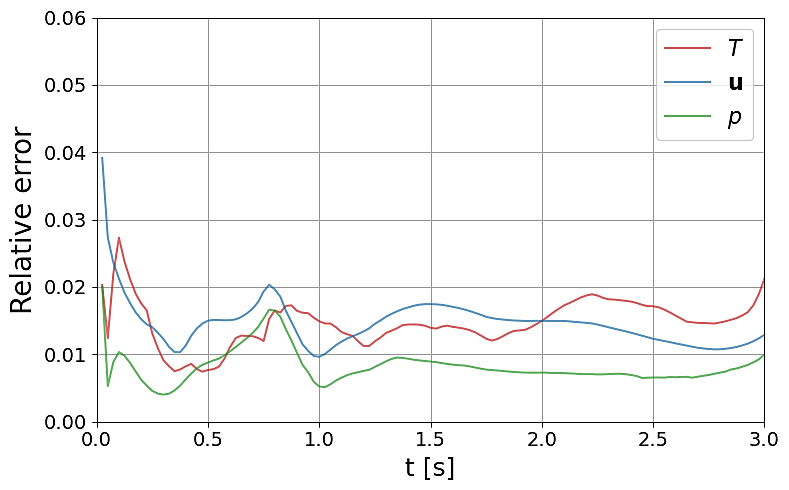}
    \caption{Temporal behavior of relative $L^2$-error of the SHRED reconstruction over time for temperature, velocity and pressure for the test case with $B_x = 1.6 \ T$ and $B_y=0.45 \, \text{T}$.}
    \label{fig: BxBy_error}
\end{figure}

Following the same procedure of the previous analysis, the relative $L^2$-error related to the SHRED reconstruction in the entire spatial domain over time has been calculated with the formula \ref{eq: error_SHRED3D}. Figure \ref{fig: BxBy_error} shows the evolution of the relative error for the temperature, velocity, and pressure fields. It can be observed that, also in this case, the errors associated with all three fields remain very low, reaching at most a few percentage points. In particular, the error trends closely resemble those observed in the case with a constant toroidal magnetic field component. The temperature error exhibits a slight increase over time, while remaining below approximately $3\%$. As explained before, this behaviour is mainly due to the initially uniform temperature field imposed by the initial condition, which progressively evolves as a consequence of the cooling effect induced by the tube located at the centre of the domain. The velocity error is initially slightly higher than the other fields, reaching values around $4\%$. However, as the flow develops and approaches a fully developed regime, the error rapidly decreases and stabilizes below $2\%$. Finally, the pressure error remains consistently low throughout the entire time window, staying below $2\%$. As discussed previously, this is expected since pressure exhibits weaker temporal fluctuations compared to velocity and temperature, making it inherently easier to reconstruct accurately.

\subsection{Time-varying magnetic field in toroidal directions}

In the previous analyses, the considered scenarios were characterized by magnetic fields that were constant in time. The present investigation extends this framework by assessing whether a parametric SHRED model can be accurate when the parameter of interest is time-dependent, specifically in the presence of temporally oscillating magnetic fields. This aspect is particularly relevant in practical MHD applications, as, in realistic operating conditions, it is extremely challenging to maintain perfectly steady magnetic field profiles. Temporal fluctuations and oscillations may commonly occur, leading to transient distortions of the desired magnetic configuration. For simplicity, only the dominant component of the magnetic field, namely the toroidal one, was considered. The temporal variations are simulated using sinusoidal profiles, allowing a controlled yet representative assessment of the model capabilities in the presence of time-dependent magnetic field fluctuations. The assumed profile is the following:
\begin{equation}
    B(t) = A \sin (\omega t + \phi) + C
\end{equation}

Consequently, snapshots have been generated by varying the values of the considered parameters $A, \omega, \phi$, and $C$. Table~\ref{tab: Bxvar_range} reports the simulated values, together with the corresponding subdivision of the data into training, validation, and test sets. Table \ref{tab: Bxvar_range} shows that three different test cases were selected, hereafter referred to as Case A, Case B, and Case C. These cases were specifically chosen because they are characterized by markedly different magnetic field profiles. 

\renewcommand{\arraystretch}{1.2}
\begin{table}[htbp]
\centering
\caption{Magnetic field configurations and subdivision of snapshots into training, validation and test sets.}
\label{tab: Bxvar_range}
\begin{tabular}{|c|c|c|c|c|}
\hline
\textbf{Set} & $A \, [\text{T}]$ & $\omega  \, [1/\text{s}] $ & $\phi \, [\text{rad}]$ & $C \, [\text{T}]$ \\ \hline
\multirow{9}{*} {\textbf{Training}}
& $0.5$ & $2 \pi / 1.4$ & $-0.05 + \pi / 2 $ & $1.3$ \\ \cline{2-5}
& $0.45$ & $2 \pi / 1.5$ & $ 1.0 + \pi / 2 $ & $1.2$ \\ \cline{2-5}
& $0.42$ & $2 \pi / 0.7$ & $ \pi / 2 + 3 $ & $1.3$ \\ \cline{2-5}
& $0.48$ & $2 \pi / 1.1$ & $ \pi / 2 + 1.3 $ & $1.1$ \\ \cline{2-5}
& $0.6$ & $2 \pi / 1.0$ & $ -1.0 $ & $1.25$ \\ \cline{2-5}
& $0.7$ & $2 \pi / 1.1$ & $ -1.0 $ & $1.1$ \\ \cline{2-5}
& $0.45$ & $2 \pi / 0.8$ & $ -0.11 $ & $1.25$ \\ \cline{2-5}
& $0.45$ & $2 \pi / 1.4$ & $ -0.11 $ & $1.25$ \\ \cline{2-5}
& $0.45$ & $2 \pi / 1.75$ & $ -0.11 $ & $1.25$ \\ \cline{2-5}
\hline
\multirow{4}{*} {\textbf{Validation}}
& $0.6$ & $2 \pi / 2.0$ & $ \pi / 2 $ & $1.1$ \\ \cline{2-5}
& $0.55$ & $2 \pi / 1.4$ & $ \pi / 2 + 1/5$ & $1.2$ \\ \cline{2-5}
& $0.45$ & $2 \pi / 1.0$ & $ -0.11 $ & $1.25$ \\ \cline{2-5}
& $0.45$ & $2 \pi / 1.5$ & $ -0.11 $ & $1.25$ \\ \cline{2-5}
\hline
\multirow{3}{*} {\textbf{Test}}
& $0.5$ & $2 \pi / 0.8$ & $ \pi / 2 $ & $1.2$ \\ \cline{2-5}
& $0.45$ & $2 \pi / 1.25$ & $ -0.11 $ & $1.25$ \\ \cline{2-5}
& $0.4$ & $2 \pi / 1.4$ & $ -0.8 $ & $1.25$ \\ \cline{2-5}
\hline
\end{tabular}
\end{table}

\begin{figure}[htbp]
    \centering
    \includegraphics[width=1.0\linewidth]{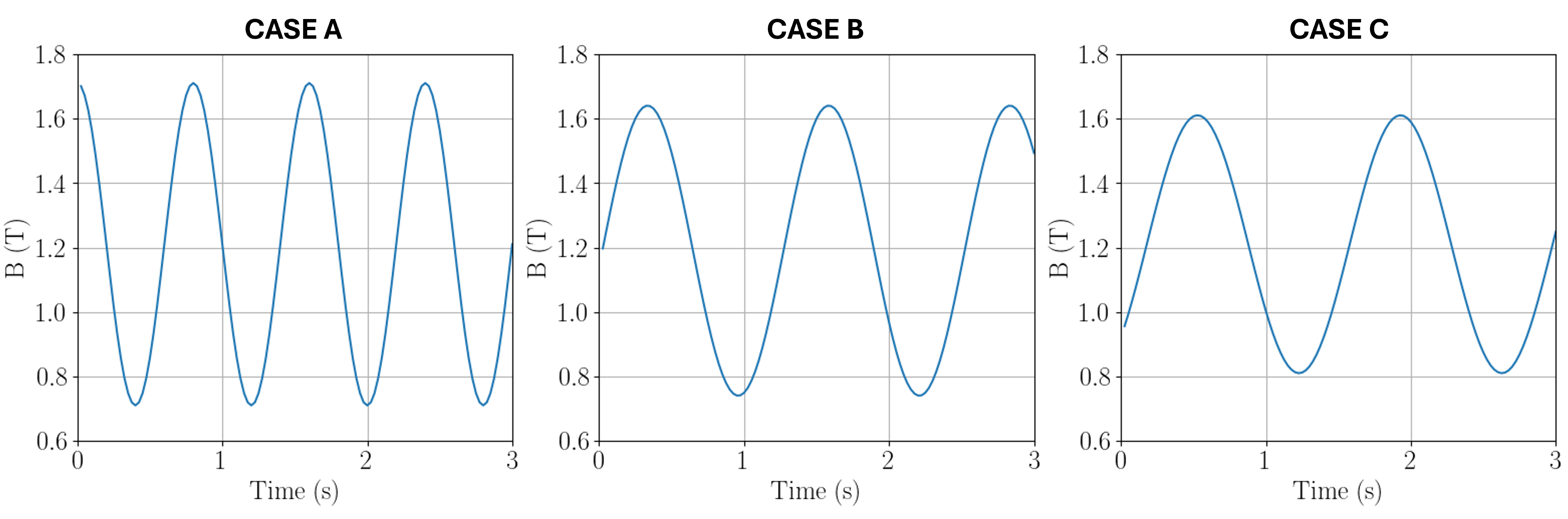}
    \caption{Temporal profiles of magnetic field of the three selected test cases.}
    \label{fig: Bxvar_profiles}
\end{figure}

Figure \ref{fig: Bxvar_profiles} illustrates the temporal evolution of the magnetic field for the three test cases. As can be observed, the profiles exhibit significantly different oscillatory behaviour, with distinct frequencies, initial values, and amplitudes. This variability allows for an assessment of the SHRED capabilities to reconstruct the system dynamics under markedly different time-dependent magnetic excitations, thereby evaluating the generality and robustness of the method for oscillating magnetic field configurations. All three test cases correspond to configurations unseen during the training phase.

\begin{figure}[htbp]
    \centering
    \includegraphics[width=1.0\linewidth]{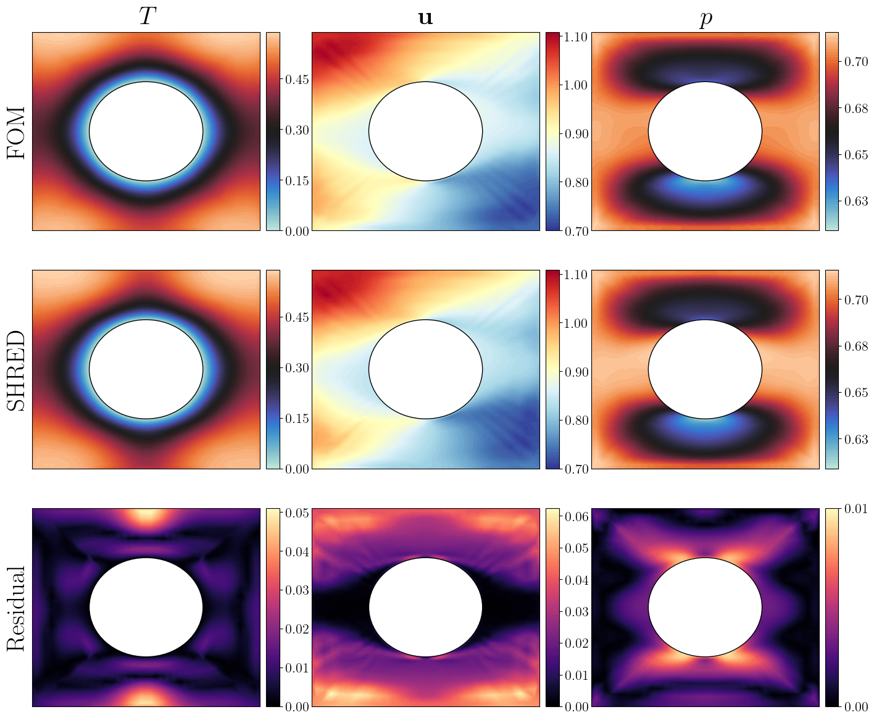}
    \caption{Results for the temperature (first column), velocity (second column) and pressure (third column) for test case A. The first row displays the reference full-order solution while the second row shows the reconstruction produced by the SHRED model. The third row reports the absolute difference between the FOM and the SHRED.}
    \label{fig: Bxvar_A}
\end{figure}
\begin{figure}[htbp]
    \centering
    \includegraphics[width=1.0\linewidth]{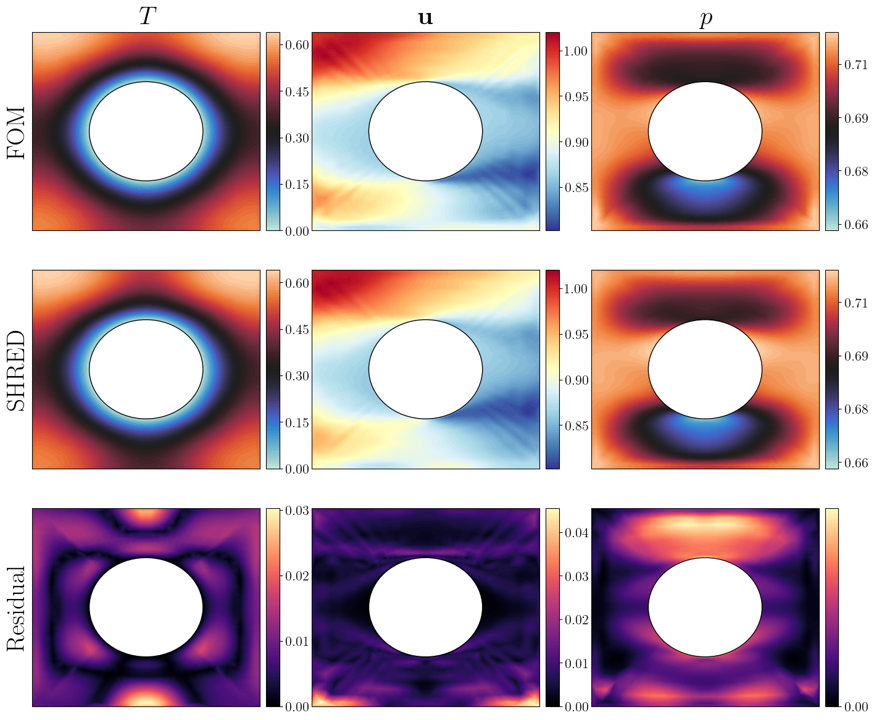}
    \caption{Results for the temperature (first column), velocity (second column) and pressure (third column) for test case B. The first row displays the reference full-order solution while the second row shows the reconstruction produced by the SHRED model. The third row reports the absolute difference between the FOM and the SHRED.}
    \label{fig: Bxvar_B}
\end{figure}
\begin{figure}[htbp]
    \centering
    \includegraphics[width=1.0\linewidth]{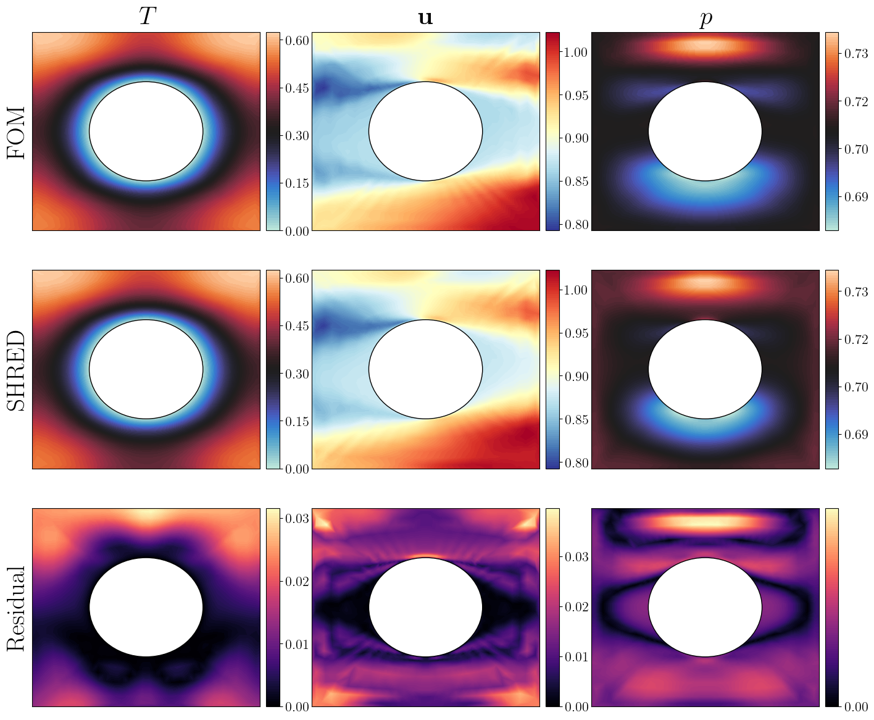}
    \caption{Results for the temperature (first column), velocity (second column) and pressure (third column) for test case C. The first row displays the reference full-order solution while the second row shows the reconstruction produced by the SHRED model. The third row reports the absolute difference between the FOM and the SHRED.}
    \label{fig: Bxvar_C}
\end{figure}

\begin{figure}[htbp]
    \centering
    \includegraphics[width=1\linewidth]{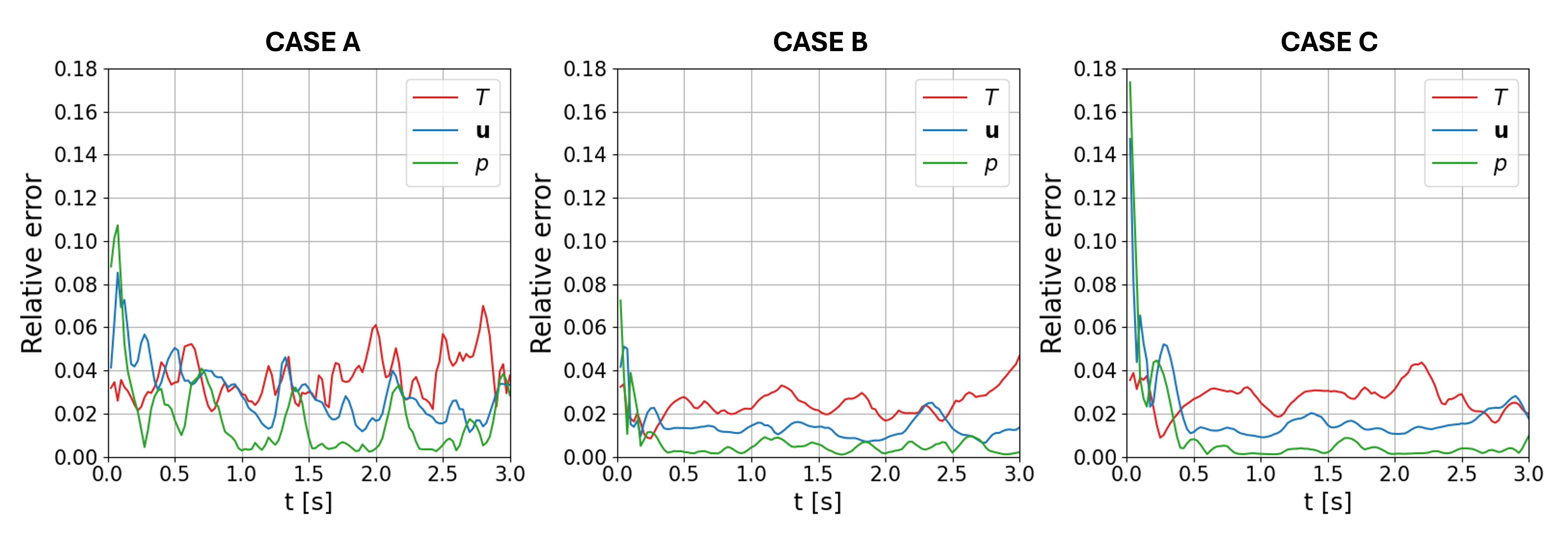}
    \caption{Temporal behavior of relative $L^2$-error of the SHRED reconstruction over time for temperature, velocity and pressure for the three test cases}
    \label{fig: Bxvar_error}
\end{figure}

Figures \ref{fig: Bxvar_A}, \ref{fig: Bxvar_B}, and \ref{fig: Bxvar_C} show the SHRED reconstruction obtained for the three test cases at the cross-section at mid-length of the geometry at $t=1.75 \ s$. It is worth noting that the flow obtained under an oscillating magnetic field is substantially different from the one observed in the case of a constant toroidal magnetic field. Moreover, by comparing the three figures, it can be observed that different temporal evolutions of the magnetic field induce substantially different effects on the system dynamics. Indeed, although the snapshots are taken at the same physical time instant $t=1.75 \ s$, the temperature, velocity, and pressure fields exhibit markedly different values and spatial distributions across the three cases. Furthermore, since the magnetic field continues to vary over time, all physical fields remain strongly time-dependent and do not converge toward a steady or quasi-steady spatial configuration. For these reasons, these scenarios represent particularly challenging test cases for state reconstruction, both because of their pronounced differences from one another and because of their highly unsteady nature. Nevertheless, it can be observed that in all three cases the trained SHRED model is able to accurately reconstruct all the considered fields, yielding residuals that are low (at most a few percentage points) and spatially localized.
Also in this case, the relative $L^2$-error of the SHRED reconstruction in the entire spatial domain over time has been calculated with Equation \ref{eq: error_SHRED3D}. Figure \ref{fig: Bxvar_error} shows the evolution of the relative error for the temperature, velocity, and pressure fields in the three cases. 
The relative errors associated with velocity and pressure start from higher initial values compared to the previously analyzed cases, especially in test case C (where they start from approximately $18\%$ and $14\%$ respectively). However, in all cases, they rapidly decrease and settle around values of $2\%$ for both fields. This behavior can be attributed to the fact that SHRED was trained on scenarios characterized by significantly different temporal magnetic field profiles. As a consequence, during the initial transient, the flow evolution strongly depends on the specific magnetic field shape imposed, leading to markedly different initial dynamics across the training cases. During the testing phase, at the very first time instants, SHRED has not yet received a sufficient number of temperature measurements to accurately infer the underlying evolution. As a result, in the three cases, the initial state reconstruction is effectively an interpolation of the initial states observed during training. Therefore, the reconstruction accuracy during the very early time instants primarily depends on how close the specific test case is to the interpolated initial estimate, rather than on the actual capability of the model to infer the underlying dynamics. As time progresses and additional temporal measurements become available, the model is able to better identify the ongoing dynamical evolution, leading to a substantial improvement in reconstruction accuracy. Instead, the temperature field starts from an already low error (less than $5\%)$, since it remains very similar across the different cases at the beginning of the simulations. Indeed, the temperature is initially homogeneous and evolves more slowly compared to velocity and pressure, requiring a longer time to develop significant spatial gradients. As the simulation progresses, the temperature errors slightly increases due to the gradual spatial diffusion driven by the cooling effect of the tube, but it remains limited, reaching a maximum of approximately between $4\%$ and $6 \%$ in the three cases. Overall, these results indicate that for all the considered physical fields, once the very early transient has elapsed, the reconstruction error remains consistently low, confirming the high accuracy of the SHRED state estimation. 

\begin{figure}[htbp]
    \centering
    \includegraphics[width=1\linewidth]{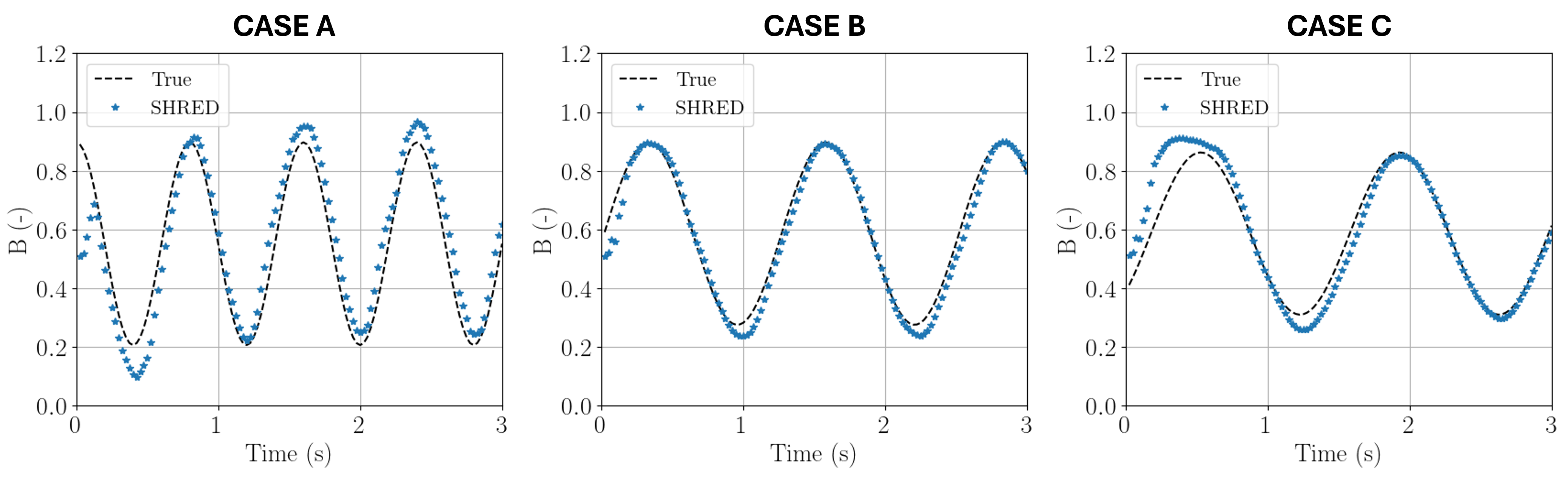}
    \caption{Real temporal profile and SHRED estimation of the rescaled magnetic field in the three test cases.}
    \label{fig: Bxvar_Bestimation}
\end{figure}

Time-dependent parameters provide an opportunity to assess whether SHRED is able not only to reconstruct the state of the system, but also to infer the parameter driving the observed dynamics. Therefore, beyond reconstructing the temperature, velocity, and pressure fields, this analysis investigates whether the model can indirectly identify and recover the time-varying magnetic field responsible for the observed MHD behavior in the three test cases. Accordingly, in this analysis, during the training phase, SHRED is provided not only with the temperature measurements and the SVD coefficients of the different state variables, but also with the data related to magnetic fields over time for the training and validation snapshots. The temporal profiles of the magnetic field were also rescaled using the same min–max normalization applied to the rest of the dataset. In this way, SHRED learns a mapping between the temperature measurements and both the state representation and the time-dependent rescaled magnetic field. As a result, the model is trained to associate observed thermal dynamics with the underlying magnetic field, enabling the inference of the magnetic field evolution from temperature data alone. \newline
Figure \ref{fig: Bxvar_Bestimation} compares the true temporal evolution of the magnetic field with the parameter estimate provided by SHRED. 

In all three cases, the estimated profile initially deviates from the reference solution: this behaviour is consistent with what was discussed previously and is related to the intrinsic nature of the LSTM recurrent block, which requires several time steps to accumulate sufficient temporal information before producing accurate inferences. At early times, the model relies on an initial interpolation of the parameter values encountered during the training phase. For the parameter as well, then, the estimation accuracy during the very early time instants depends on how close the specific profile starts to the interpolated initial estimate. However, after this transient phase, in all three test cases, the estimated magnetic field closely follows the true profile, providing an accurate reconstruction for the remainder of the time window. These results therefore demonstrate that the SHRED model, when trained in this manner, is also capable of providing an accurate estimate of the magnetic field evolution for test cases characterized by significantly different magnetic profiles. This confirms the robustness of the approach in reconstructing, in a general and reliable manner, oscillating magnetic field dynamics. Then SHRED could also serve as a potential diagnostic tool for fusion reactors. By exploiting measurements within the blanket, the method may be used to estimate the magnetic field impinging on it, thus offering indirect insight into the temporal evolution of magnetic field conditions in neighboring regions of the reactor.

\section{Conclusions}\label{sec: Conclusions}
This study investigated the application of the SHallow REcurrent Decoder (SHRED) to magnetohydrodynamic flows relevant to liquid-metal blankets in nuclear fusion reactors. SHRED was applied to a three-dimensional configuration representative of a portion of a WCLL blanket cell, specifically the flow of lead-lithium around a cold tube modelling a water-cooled pipe. This work constitutes the first application of SHRED to a geometry directly related to a liquid-metal blanket configuration. The main objective was to evaluate the capability of SHRED to accurately reconstruct the complete system state relying exclusively on temperature measurements acquired at three randomly distributed sensors. Several magnetic field configurations were examined, and separate SHRED models were trained and tested for each configuration. The analysis first focused on a scenario with a constant toroidal magnetic field, representing the dominant magnetic field component experienced within the blanket. When trained on cases spanning different toroidal magnetic field intensities, SHRED was shown to accurately reconstruct the system state both for test cases lying within the training range and for a configuration outside the training interval. The complexity of the problem was then increased by introducing an additional poloidal magnetic field component. Also in this more complex setting, SHRED demonstrated high reconstruction accuracy for unseen magnetic field intensities and orientations. Finally, a configuration characterized by a time-varying toroidal magnetic field was investigated. Even in this strongly time-dependent scenario, SHRED proved capable of accurately reconstructing the temporal evolution of the system state despite the highly dynamic nature of the flow. This case is particularly relevant from a practical perspective, as maintaining perfectly steady magnetic field profiles over time is extremely challenging under realistic fusion reactor operating conditions, and temporal fluctuations are expected to occur. Moreover, in this final configuration, the ability of SHRED to infer the magnetic field itself (using only temperature measurements) was also assessed. The results indicate that SHRED can accurately estimate the temporal evolution of the magnetic field parameter, further underscoring the robustness and versatility of the proposed methodology. These findings suggest that SHRED may serve not only as a state reconstruction tool but also as a diagnostic instrument in fusion reactors. Indeed, by leveraging measurements of a single blanket observable, the framework can also infer the magnetic field interacting with the blanket, thereby potentially providing indirect information on the evolution of magnetic field conditions in adjacent regions, such as the plasma chamber. \newline
Overall, the proposed methodology constitutes a computationally efficient and fully data-driven framework for MHD state reconstruction in real-time or multi-query settings. SHRED demonstrated a consistently high level of accuracy across a wide range of different magnetic field configurations. So SHRED has strong potential for accurate state reconstruction under general magnetic field conditions, encompassing variations in intensity, spatial orientation, and temporal behavior. The presented analysis demonstrate that SHRED may constitute a highly effective and precise AI-based model for reconstructing MHD flows in blanket configurations. The approach requires measurements of only a single physical field (for example, temperature, which is the easiest to measure) at a limited number of spatial locations, enabling flexible sensor placement within the reactor. In addition, the training phase can be completed in only a few minutes on standard laptops, thanks to the training on the SVD reduced level. Finally, the promising results obtained with SHRED open the door to several future research avenues. These include the application of SHRED to entire WCLL cell geometries, the integration of state estimation into closed-loop control strategies, and the deployment of these models in digital-twin frameworks for fusion reactors. In this perspective, the methodologies developed in this study represent a solid foundation for future research aimed at enabling real-time monitoring, diagnostics, and control of liquid-metal blankets and represent a significant step toward efficient, predictive and operationally relevant modelling of MHD physics in next-generation fusion reactors.

\clearpage

\bibliography{bibliography.bib}

@book{freidberg2014ideal,
  title={Ideal MHD},
  author={Freidberg, Jeffrey P},
  year={2014},
  publisher={Cambridge University Press}
}

@article{molokov2007liquid,
  title={Liquid metal magnetohydrodynamics for fusion blankets},
  author={Molokov, Sergei and Moreau, Ren{\'e} and Moffatt, Keith and B{\"u}hler, Leo},
  journal={Magnetohydrodynamics: Historical Evolution and Trends},
  pages={171--194},
  year={2007},
  publisher={Springer}
}

@article{ferrero2023impact,
  title={Impact Assessment of Radiative Heat Transport in ARC-Class Reactor FLiBe Liquid Immersion Blanket},
  author={Ferrero, Gabriele and Testoni, Raffaella and Zucchetti, Massimo},
  journal={Nuclear Science and Engineering},
  pages={1--16},
  year={2023},
  publisher={Taylor \& Francis}
}

@book{biskamp1997nonlinear,
  title={Nonlinear magnetohydrodynamics},
  author={Biskamp, Dieter},
  edition={1},
  year={1997},
  publisher={Cambridge University Press}
}

@incollection{buhler2007liquid,
  title={Liquid metal magnetohydrodynamics for fusion blankets},
  author={B{\"u}hler, Leo},
  booktitle={Magnetohydrodynamics: Historical Evolution and Trends},
  pages={171--194},
  year={2007},
  publisher={Springer}
}

@book{muller2001magnetofluiddynamics,
  title={Magnetofluiddynamics in channels and containers},
  author={M{\"u}ller, Ulrich and B{\"u}hler, Leo},
  year={2001},
  publisher={Springer Science \& Business Media}
}

@article{lassila2014model,
  title={Model order reduction in fluid dynamics: challenges and perspectives},
  author={Lassila, Toni and Manzoni, Andrea and Quarteroni, Alfio and Rozza, Gianluigi},
  journal={Reduced Order Methods for modeling and computational reduction},
  pages={235--273},
  year={2014},
  publisher={Springer}
}

@book{rozza_model_2020,
	title = {Model Order Reduction: Volume 2: Snapshot-Based Methods and Algorithms},
	publisher = {De Gruyter},
	author = {Rozza, Gianluigi and Hess, Martin and Stabile, Giovanni and Tezzele, Marco and Ballarin, Francesco and Gräßle, Carmen and Hinze, Michael and Volkwein, Stefan and Chinesta, Francisco and Ladeveze, Pierre and Maday, Yvon and Patera, Anthony and Farhat Char, J},
	year = {2020},
}

@article{loverso2024application,
  title={Application of a non-intrusive reduced order modeling approach to magnetohydrodynamics},
  author={Lo Verso, M and Riva, S and Introini, C and Cervi, E and Giacobbo, F and Savoldi, L and Di Prinzio, M and Caramello, M and Barucca, L and Cammi, A},
  journal={Physics of Fluids},
  volume={36},
  number={10},
  year={2024},
  publisher={AIP Publishing}
}

@article{riva2025data,
  title={Data-driven reduced order modelling with malfunctioning sensors recovery applied to the Molten Salt Reactor case},
  author={Riva, Stefano and Introini, Carolina and Zio, Enrico and Cammi, Antonio},
  journal={EPJ Nuclear Sciences \& Technologies},
  volume={11},
  pages={55},
  year={2025},
  publisher={EDP Sciences}
}

@article{riva2025real,
  title={Real-Time State Estimation of Neutron Flux in Molten Salt Fast Reactors from Out-Core Sparse Measurements},
  author={Riva, Stefano and Deanesi, Sophie and Introini, Carolina and Lorenzi, Stefano and Cammi, Antonio},
  journal={Nuclear Science and Engineering},
  pages={1--14},
  year={2025},
  publisher={Taylor \& Francis}
}

@inproceedings{riva2024impact,
  title={Impact of Malfunctioning Sensors on Data-Driven Reduced Order Modelling: Application to Molten Salt Reactors},
  author={Riva, Stefano and Introini, Carolina and Zio, Enrico and Cammi, Antonio},
  booktitle={EPJ Web of Conferences},
  volume={302},
  pages={17003},
  year={2024},
  organization={EDP Sciences}
}

@article{riva2024multi,
  title={Multi-physics model bias correction with data-driven reduced order techniques: Application to nuclear case studies},
  author={Riva, Stefano and Introini, Carolina and Cammi, Antonio},
  journal={Applied Mathematical Modelling},
  volume={135},
  pages={243--268},
  year={2024},
  publisher={Elsevier}
}

@article{cammi2024data,
  title={Data-driven model order reduction for sensor positioning and indirect reconstruction with noisy data: Application to a Circulating Fuel Reactor},
  author={Cammi, Antonio and Riva, Stefano and Introini, Carolina and Loi, Lorenzo and Padovani, Enrico},
  journal={Nuclear Engineering and Design},
  volume={421},
  pages={113105},
  year={2024},
  publisher={Elsevier}
}

@article{RoyKutz2018_Plasma,
    author = {Taylor, Roy and Kutz, J. Nathan and Morgan, Kyle and Nelson, Brian A.},
    title = "{Dynamic mode decomposition for plasma diagnostics and validation}",
    journal = {Review of Scientific Instruments},
    volume = {89},
    number = {5},
    pages = {053501},
    year = {2018},
    month = {05}
}

@article{kaptanoglu2021physics,
  title={Physics-constrained, low-dimensional models for magnetohydrodynamics: First-principles and data-driven approaches},
  author={Kaptanoglu, Alan A and Morgan, Kyle D and Hansen, Chris J and Brunton, Steven L},
  journal={Physical Review E},
  volume={104},
  number={1},
  pages={015206},
  year={2021},
  publisher={APS}
}

@article{kaptanoglu2020characterizing,
  title={Characterizing magnetized plasmas with dynamic mode decomposition},
  author={Kaptanoglu, Alan A and Morgan, Kyle D and Hansen, Chris J and Brunton, Steven L},
  journal={Physics of Plasmas},
  volume={27},
  number={3},
  year={2020},
  publisher={AIP Publishing}
}

@inproceedings{loverso2024solver,
    author = {Lo Verso, Matteo and Introini, Carolina and Cervi, Eric and Giacobbo, Francesca and Cammi, Antonio },
    title = {A novel Openfoam library for magneto-hydrodynamics studies in the nuclear fusion field},
    booktitle = {Proceedings of the NUTHOS-14 International Conference, Vancouver (CAN)},
    year = {2024},
}

@article{seo2024avoiding,
  title={Avoiding fusion plasma tearing instability with deep reinforcement learning},
  author={Seo, Jaemin and Kim, SangKyeun and Jalalvand, Azarakhsh and Conlin, Rory and Rothstein, Andrew and Abbate, Joseph and Erickson, Keith and Wai, Josiah and Shousha, Ricardo and Kolemen, Egemen},
  journal={Nature},
  volume={626},
  number={8000},
  pages={746--751},
  year={2024},
  publisher={Nature Publishing Group UK London}
}

@article{jalalvand2021real,
  title={Real-time and adaptive reservoir computing with application to profile prediction in fusion plasma},
  author={Jalalvand, Azarakhsh and Abbate, Joseph and Conlin, Rory and Verdoolaege, Geert and Kolemen, Egemen},
  journal={IEEE Transactions on Neural Networks and Learning Systems},
  volume={33},
  number={6},
  pages={2630--2641},
  year={2021},
  publisher={IEEE}
}

@article{degrave2022magnetic,
  title={Magnetic control of tokamak plasmas through deep reinforcement learning},
  author={Degrave, Jonas and Felici, Federico and Buchli, Jonas and Neunert, Michael and Tracey, Brendan and Carpanese, Francesco and Ewalds, Timo and Hafner, Roland and Abdolmaleki, Abbas and de Las Casas, Diego and others},
  journal={Nature},
  volume={602},
  number={7897},
  pages={414--419},
  year={2022},
  publisher={Nature Publishing Group UK London}
}

@article{wang2025learning,
  title={Learning plasma dynamics and robust rampdown trajectories with predict-first experiments at TCV},
  author={Wang, Allen M and Pau, Alessandro and Rea, Cristina and So, Oswin and Dawson, Charles and Sauter, Olivier and Boyer, Mark D and Vu, Anna and Galperti, Cristian and Fan, Chuchu and others},
  journal={Nature Communications},
  volume={16},
  number={1},
  pages={8877},
  year={2025},
  publisher={Nature Publishing Group UK London}
}

@article{battye2025digital,
  title={Digital Twins in Fusion Energy Research: Current State and Future Directions},
  author={Battye, Michael I and Perinpanayagam, Suresh},
  journal={IEEE Access},
  year={2025},
  publisher={IEEE}
}

@article{LOVERSO2025115080,
title = {Enhancing computational efficiency in nuclear fusion through reduced order modelling: Applications in magnetohydrodynamics},
journal = {Fusion Engineering and Design},
volume = {216},
pages = {115080},
year = {2025},
issn = {0920-3796},
doi = {https://doi.org/10.1016/j.fusengdes.2025.115080},
url = {https://www.sciencedirect.com/science/article/pii/S0920379625002777},
author = {Matteo {Lo Verso} and Stefano Riva and Carolina Introini and Eric Cervi and Luciana Barucca and Marco Caramello and Matteo {Di Prinzio} and Francesca Giacobbo and Laura Savoldi and Antonio Cammi},
}

@article{williams2024sensing,
  title={Sensing with shallow recurrent decoder networks},
  author={Williams, Jan P and Zahn, Olivia and Kutz, J Nathan},
  journal={Proceedings of the Royal Society A},
  volume={480},
  number={2298},
  pages={20240054},
  year={2024},
  publisher={The Royal Society}
}

@article{LSTM2016long,
  title={Long short-term memory},
  author={Computation, Neural},
  journal={Neural Comput},
  volume={9},
  pages={1735--1780},
  year={2016}
}

@article{SDN20shallow,
  title={Shallow neural networks for fluid flow reconstruction with limited sensors},
  author={Erichson, N Benjamin and Mathelin, Lionel and Yao, Zhewei and Brunton, Steven L and Mahoney, Michael W and Kutz, J Nathan},
  journal={Proceedings of the Royal Society A},
  volume={476},
  number={2238},
  pages={20200097},
  year={2020},
  publisher={The Royal Society Publishing}
}

@article{faraji2025shallow,
  title={Shallow recurrent decoder for reduced order modeling of E$\times$ B plasma dynamics},
  author={Faraji, Farbod and Reza, Maryam and Kutz, J Nathan},
  journal={Machine Learning: Science and Technology},
  volume={6},
  number={2},
  pages={025024},
  year={2025},
  publisher={IOP Publishing}
}

@article{tomasetto2025reduced,
  title={Reduced order modeling with shallow recurrent decoder networks},
  author={Tomasetto, Matteo and Williams, Jan P and Braghin, Francesco and Manzoni, Andrea and Kutz, J Nathan},
  journal={arXiv preprint arXiv:},
  year={2025}
}

@article{loverso2026SHRED,
  title={Reduced Order Modelling for nuclear fusion with parametric Shallow Recurrent Decoder Networks: Applications to Magnetohydrodynamics},
  author={Lo Verso, Matteo and Introini, Carolina and Cammi, Antonio and Kutz, J Nathan},
  journal={arXiv preprint arXiv:2502.10930},
  year={2026}
}

@inproceedings{takens2006detecting,
  title={Detecting strange attractors in turbulence},
  author={Takens, Floris},
  booktitle={Dynamical Systems and Turbulence, Warwick 1980: proceedings of a symposium held at the University of Warwick 1979/80},
  pages={366--381},
  year={2006},
  organization={Springer}
}

@article{riva2025constrained,
  title={Constrained Sensing and Reliable State Estimation with Shallow Recurrent Decoders on a TRIGA Mark II Reactor},
  author={Riva, Stefano and Introini, Carolina and Kutz, Jos{\`e} Nathan and Cammi, Antonio},
  journal={arXiv preprint arXiv:2510.12368},
  year={2025}
}

@article{introini2025models,
  title={From Models To Experiments: Shallow Recurrent Decoder Networks on the DYNASTY Experimental Facility},
  author={Introini, Carolina and Riva, Stefano and Kutz, J Nathan and Cammi, Antonio},
  journal={arXiv preprint arXiv:2503.08907},
  year={2025}
}

@article{riva2025towards,
  title={Towards Efficient Parametric State Estimation in Circulating Fuel Reactors with Shallow Recurrent Decoder Networks},
  author={Riva, Stefano and Introini, Carolina and Kutz, J Nathan and Cammi, Antonio},
  journal={arXiv preprint arXiv:2503.08904},
  year={2025}
}

@article{riva2025robust,
  title={Robust state estimation from partial out-core measurements with shallow recurrent decoder for nuclear reactors},
  author={Riva, Stefano and Introini, Carolina and Cammi, Antonio and Kutz, J Nathan},
  journal={Progress in Nuclear Energy},
  volume={189},
  pages={105928},
  year={2025},
  publisher={Elsevier}
}

@article{moen2025mapping,
  title={Mapping surface height dynamics to subsurface flow physics in free-surface turbulent flow using a shallow recurrent decoder},
  author={Moen, Kristoffer S and Aarnes, J{\o}rgen R and Ellingsen, Simen {\AA} and Kutz, J Nathan},
  journal={arXiv preprint arXiv:2510.06202},
  year={2025}
}

@inproceedings{loverso_NURETH,
    author = {Lo Verso, Matteo and Riva, Stefano and Introini, Carolina and Cervi, Eric and Giacobbo, Francesca and Di Prinzio, Matteo and Caramello, Marco and Savoldi, Laura and Cammi, Antonio },
    title = {A Novel Parametric Dynamic Mode Decomposition Formulation: Application to Magnetohydrodynamic Liquid Metal Flows},
    booktitle = {Proceedings of the NURETH-21 International Conference, Busan (Korea)},
    year = {2025},
}

@article{arena2021demo,
  title={The DEMO water-cooled lead--lithium breeding blanket: design status at the end of the pre-conceptual design phase},
  author={Arena, Pietro and Del Nevo, Alessandro and Moro, Fabio and Noce, Simone and Mozzillo, Rocco and Imbriani, Vito and Giannetti, Fabio and Edemetti, Francesco and Froio, Antonio and Savoldi, Laura and others},
  journal={Applied Sciences},
  volume={11},
  number={24},
  pages={11592},
  year={2021},
  publisher={MDPI}
}

@article{tassone2022magnetic,
  title={On the magnetic field distribution in the TBM set and a blanket based on the DEMO2017 baseline},
  author={Tassone, Alessandro},
  year={2022}
}

@article{lyu20253d,
  title={3D magneto-convective instabilities of liquid metal flow in a generic geometry related to WCLL blankets},
  author={Lyu, B and B{\"u}hler, L and Koehly, C and Mistrangelo, C},
  journal={Fusion Engineering and Design},
  volume={217},
  pages={115113},
  year={2025},
  publisher={Elsevier}
}

@article{koehly2019design,
  title={Design of a test section to analyze magneto-convection effects in WCLL blankets},
  author={Koehly, C and B{\"u}hler, L and Mistrangelo, C},
  journal={Fusion Science and Technology},
  volume={75},
  number={8},
  pages={1010--1015},
  year={2019},
  publisher={Taylor \& Francis}
}

@article{buhler2024liquid,
  title={Liquid metal MHD research at KIT: Fundamental phenomena and flows in complex blanket geometries},
  author={B{\"u}hler, L and Brinkmann, H-J and Courtessole, C and Kl{\"u}ber, V and Koehly, C and Lyu, B and Mistrangelo, C and Roth, J},
  journal={Fusion Engineering and Design},
  volume={200},
  pages={114195},
  year={2024},
  publisher={Elsevier}
}

\end{document}